\documentclass[utf8]{article}

\usepackage{url,hyperref,lineno,microtype,subcaption,graphicx}
\usepackage[onehalfspacing]{setspace}
\usepackage{amssymb}
\usepackage{amsmath,amsfonts,bm}
\usepackage{wrapfig}
\usepackage{float}
\newcommand{\itcaption}[1]{\caption{\small {#1}}}
\newcommand{\itsubcaption}[1]{\caption{\small {#1}}}
\newcommand{\mb}[1]{\mathbf{#1}}
\newcommand{\mc}[1]{\mathcal{#1}}
\newcommand{\lbound}[1]{\lfloor{#1}\rfloor}
\newtheorem{theorem}{Theorem}[section]
\newtheorem{definition}[theorem]{Definition}

\DeclareMathOperator*{\argmax}{arg\,max}

\newbool{arxiv}

\booltrue{arxiv}

\ifbool{arxiv}{}
{\linenumbers}

\newcommand{\revision}[1]{{#1}}

\newcommand{\rih}[1]{{\color{magenta}RIH: #1}}
\renewcommand{\v}{\mb}
\newcommand{\calN}{{\mathcal N}}
\newcommand{\SKIP}[1]{}

\newcommand{\Ex}{E}

\newcommand*{\colorboxed}{}
\def\colorboxed#1#{%
  \colorboxedAux{#1}%
}
\newcommand*{\colorboxedAux}[3]{%
  \begingroup
    \colorlet{cb@saved}{.}%
    \color#1{#2}%
    \boxed{%
      \color{cb@saved}%
      #3%
    }%
  \endgroup
}

\def\keyFont{\fontsize{8}{11}\helveticabold }
\def\firstAuthorLast{Peter Tu {et~al.}} 
\def\Authors{Peter Tu\,$^{1,*}$, Zhaoyuan Yang\,$^{1}$, Richard
Hartley\,$^{2}$,
Zhiwei Xu\,$^{2}$, Jing Zhang\,$^{2}$, Yiwei Fu\,$^{1}$, Dylan
Campbell\,$^{2}$,
Jaskirat Singh\,$^{2}$, and Tianyu Wang\,$^{2}$}
\def\Address{$^{1}$ Computer Vision and Machine Learning Laboratory, General
Electric Research, Niskayuna, New York, USA\\
$^{2}$School of Computing, College of Engineering, Computing and Cybernetics,
Australian National University, Canberra, Australia}
\def\corrAuthor{Corresponding Author: Peter Tu}

\def\corrEmail{tu@ge.com}

\begin{document}
\onecolumn
\firstpage{1}

\title {Probabilistic and Semantic Descriptions of Image Manifolds and Their
Applications} 

\author[\firstAuthorLast]{\Authors}
\address{} 
\correspondance{} 

\extraAuth{}

\maketitle

\begin{abstract}

This paper begins with a description of methods for estimating probability
density functions for images that reflects the observation that such data is
usually constrained to lie in restricted regions of the high-dimensional image
space --- not every pattern of pixels is an image. It is common to say that
images lie on a lower-dimensional manifold 
 in the high-dimensional space.
However, although images may lie on such lower-dimensional manifolds, it is
not
the case that all points on the manifold have an equal probability of being
images. Images are unevenly distributed on the manifold, and our task is to
devise ways to model this distribution as a probability distribution.
In pursuing this goal, we consider generative models that are popular in AI
and
computer vision community. For our purposes, generative/probabilistic models
should have the properties of 1) sample generation: it should be possible to
sample from this distribution according to the modelled density function, and
2)
probability computation: given a previously unseen sample from the dataset of
interest, one should be able to compute the probability of the sample, at
least
up to a normalising constant. To this end, we investigate the use of methods
such as normalising flow and diffusion models.
\revision{We then show how semantic interpretations are used to describe
points on the manifold.
To achieve this,} we consider an emergent language framework that makes
use of variational encoders to produce a disentangled representation of points
that reside on a given manifold. Trajectories between points on a manifold can
then be described in terms of evolving semantic descriptions.
\revision{In addition to describing the manifold in terms of density and
semantic disentanglement, we also show that such probabilistic descriptions
(bounded) can be used to improve semantic consistency by constructing
defences against adversarial attacks.
We evaluate our methods on CelebA and point samples for likelihood estimation with improved semantic robustness and out-of-distribution detection capability,
MNIST and CelebA for semantic disentanglement with explainable and editable semantic interpolation,
and CelebA and Fashion-MNIST to defend against patch attacks with significantly improved classification accuracy.
We also discuss the limitation of applying our likelihood estimation to 2D images in diffusion models.
}

\tiny
 \keyFont{ \section{Keywords:} image manifold, normalising flow, diffusion
model, likelihood estimation, semantic
disentanglement, adversarial attacks and defences}
\end{abstract}

%
\section{Introduction}
Understanding the complex probability distribution of the data is essential
for
image authenticity and quality analysis, but is challenging due to its high
dimensionality and intricate domain variations~
\citep{manifold_dimension,manifold_geo}. Seen images usually have high
probabilities on a low-dimensional manifold embedded in the higher-dimensional
space of the image encoder.
\revision{
Nevertheless, the phenomenon that image embeddings encoded using methods such as
a pretrained CLIP encoder~\citep{dall_e} lie within a narrow cone of the unit
sphere instead of the entire sphere~\citep{cone_1,cone_2}, which degrades the
aforementioned pattern of probability distribution.
Hence, on such a manifold, it is unlikely that} every point can
be
decoded into a realistic image because of the unevenly distributed
probabilities. Therefore, it is important to compute the probability in the
latent space to indicate whether the corresponding image is in a high-density
region of the space~
\citep{manifold_mle,manifold_flow_flow,manifold_flow_gan,nf_prob,manifold_gan,manifold_hyperbolic,manifold_flow_beyond}. 
This helps to
distinguish seen images from unseen images, or synthetic images from real
images.
Some works train a discriminator with positive (real) and negative
(synthetic) examples in the manner of contrastive
learning~\citep{detect_gen_by_real} or analyse their frequency
differences~\citep{frequency}. However, they do not address this problem using
the probabilistic framework afforded by modern generative models.

In this work, we calculate the exact log-probability of an image by
utilising generative models that assign high probabilities to seen images
and low probabilities to unseen images.
\revision{The confidence of such probabilities is usually related to
image fidelity,
we hence also introduce efficient and effective (with improved semantic robustness)
generation strategies using hierarchical
structure and large sampling steps with the Runge-Kutta method
(RK4)~\citep{RK41,RK42} for stabilisation.}
Specifically, we use normalising flow
(NF)~\citep{vae_nf,nf_prob} and diffusion
models (DMs)~\citep{ddpm,ddpm_understand,ddim} as image generators. NF models
learn an image embedding space that conforms to a predefined distribution,
usually a Gaussian.
In contrast, DMs diffuse images with Gaussian noise in each
forward step and learn denoising gradients for the backward steps.
A random sample from the Gaussian distribution can then be
analytically represented on an image manifold
and visualised through an image decoder (for NF models) or denoiser (for
diffusion models).
\revision{In prior works, NF for exact likelihood estimation~\citep{vae_nf,exact_1,exact_2}
and with hierarchical structure~\citep{hir_structure_1,hir_structure_2,hir_structure_3} have
been explored in model training.
To the best of our knowledge, however, it has not been studied by investigating such
likelihood distribution of seen and unseen images with a hierarchical
structure (without losing the image quality) from the manifold perspective.
This is also applied to the diffusion models noting the
difficulty of combining such exact likelihood with the mean squared error
loss in diffusion training.}

\revision{Samples from these image generators} can be thought of having several meaningful semantic attributes.
It is often desirable that these attributes be orthogonal to each other
in the sample latent space so as to achieve a controllable and
interpretable representation.
In this work, \revision{we disentangle semantics
in the latent space by using a variational autoencoder
(VAE)~\citep{kingma2013auto} in the framework of emergent languages
(EL)~
\citep{havrylov2017emergence,el_control,el_communication,el_dynamic,el_miccai}. 
This allows the latent representation on the manifold to be more
robust, interpretable, compositional, controllable, and transferable.
\revision{Although some VAE variant models such as $\beta$-TCVAE~\citep{tcvae},
GuidedVAE~\citep{guidedvae}, and DCVAE~\citep{dcvae} achieve qualified semantic
disentanglement results,
we mainly focus on understanding the effectiveness of the emergent language
framework for VAE based disentanglement inspired by \citep{el_model} and
emphasizing the feasibility of our GridVAE (with mixture of Gaussian priors)
under such an EL framework to study semantic distributions on the image manifold.}
We also evaluate their semantic robustness on such a manifold against
adversarial and patch attacks~
\citep{patch_adv,patch_check,patch_gan,patch_guard,patch_sentinet,PGD_attack,Ensemble_Adversarial_Training_Attacks_defenses,Towards_Evaluating_Robustness_Neural_Networks} and defend against the same attacks using
semantic consistency with a purification loss.}

We organise this paper into three sections, each with their own
experiments: log-likelihood estimation for a given image under normalising
flows and diffusion models (see Section~\ref{sec:mle}),
\revision{semantic disentanglement in emergent languages
for a latent representation of object attributes, using a proposed
GridVAE model (see Section~\ref{sec:el}), and adversarial
attacks and defences in image space to preserve semantics (see
Section~\ref{sec:attacks}).}

%
\section{Likelihood Estimation with Image Generators}
\label{sec:mle}

We evaluate the log-probability of a given image using 1) a hierarchical
normalising flow model, 2) a diffusion model adapted to taking large
sampling steps,
and 3) a diffusion model that uses a higher-order solution to increase
generation robustness.

\subsection{Hierarchical Normalising Flow Models}
\label{sec:nf}

\begin{figure}[!ht]
\centering
\includegraphics[width=0.5\linewidth]{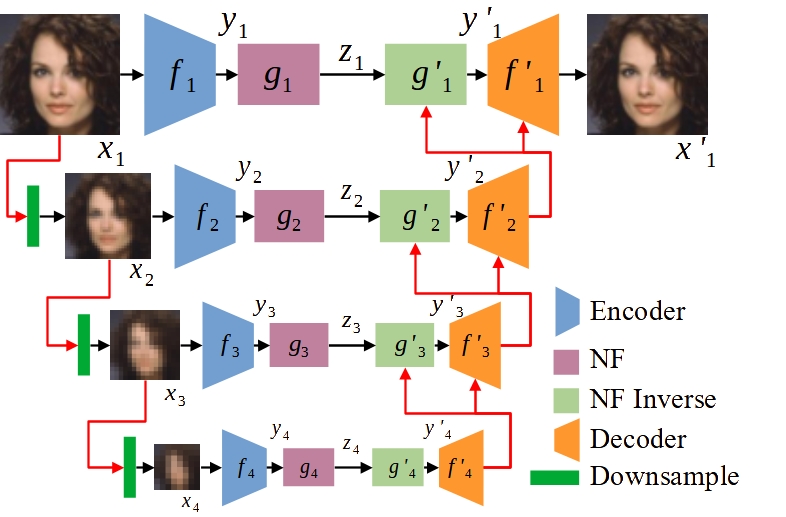}
\itcaption{
A 4-level hierarchical normalising flow model, where 
each level involves the functions $(f_i,g_i,g'_i,f'_i)$.
The normalising flow (NF) model is based on Glow~\citep{glow}; the
downsampling block decreases image resolution by a factor of two; and
the output of each higher ($i>1$)
level is conditioned on the output of the lower level.
We first train all autoencoders $\{f_i,f'_i\}$ jointly,
then train all flows $\{g_i,g'_i\}$ jointly,
to obtain the generated image $\mb{x}'_1$.
The latent variable $\mb z_i$ conforms to the standard Gaussian distribution
$\mathcal{N}(\mb{0},\mb{1})$ during training;
at test time, $\mb{z}_i$ is sampled from $\mathcal{N}(\mb{0},\mb{1})$ for image
generation.}
\label{fig:flowchart}
\end{figure}

Normalising flow (NF) refers to a sequence of invertible functions that
may be used to transform a high-dimensional image space into a low-dimensional
embedding space corresponding to a probability distribution, usually Gaussian.
Dimensionality reduction is achieved via an autoencoding framework.
For the hierarchical model, the latent vector corresponding to the image
$\mb{x}_i$ at each level $i$ is computed as
\begin{equation}
    \mb{z}_i = g_i (\mb{y}_i) = g_i \circ f_i (\mb{x}_i) \sim
\mathcal{N}(\mb{0}, \mb{1})\ ,
\end{equation}
and the inversion of this process reconstructs the latent
$\mb{z}'_i$
to $\mb{x}'_i$ as
\begin{equation}
    \mb{x}'_i = f'_i \circ g'_i (\mb{z}'_i)\ ,
\end{equation}
where the decoder $f'_i$ and flow inverse function $g'_i$ are inversions
of the encoder $f_i$
and
flow function $g$ respectively, and $\mb{z}'_i$ can be $\mb{z}_i$ or
randomly sampled
from $\mathcal{N}(\mb{0}, \mb{1})$.
We illustrate hierarchical autoencoders and flows for rich and high-level
spatial
information with conditioning variables in either image space or latent space.
In Fig.~\ref{fig:flowchart}, we show a 4-level hierarchical normalising
flow model, where each set of functions ($f_i,g_i,g'_i,f'_i$) corresponds
to one level and where
$g'_i$ and $f'_i$ are conditioned on the higher-level
reconstruction,
that is
\begin{equation}
    \mb{x}'_1 = f'_1 \circ g'_1 (\mb{z}'_1 \vert f'_2 \circ g'_2
(\mb{z}'_2 \vert f'_3 \circ g'_3 (\mb{z}'_3 \vert f'_4 \circ g'_4
(\mb{z}'_4))))\ .
\end{equation}

The model is learned in two phases: joint learning of all
autoencoders
$\{f_i,f'_i\}$ and then joint learning of all flows $\{g_i,g'_i\}$ with the
pretrained autoencoders, for all $i \in \{1, 2, 3, 4\}$.
The loss function for autoencoder learning, denoted
$\mc{L}_{\text{ae}}$,
is the mean squared error (MSE)
between the reconstructed data
and the processed data, and for the learning of flows the objective is to
minimise
the negative log-probability
of $\mb{y}_i$,
denoted $\mc{L}_{\text{flow}}$, such that the
represented distribution of the latent variable is modelled to be the standard
Gaussian distribution, from which a random latent variable can be sampled for
data generation. Given $N$ pixels and $C$ channels ($C=3$ for an RGB image
and $C=1$ for a greyscale image), $\mb{x}_i$ at level $i$ can be represented as
$\mb{x}_i = \{\mb{x}_{ij}\}$ for all $j \in \{1, ..., N\}$,
the autoencoder loss is then given by
\begin{equation}
    \mathcal{L}_{\text{ae}} (\mb{x}'_{i}, \mb{x}_{i})
    = \frac{1}{CN} \sum^N_{j=1} \| \mb{x}'_{ij} - \mb{x}_{ij} \|^2\ ,
\end{equation}
and the flow loss for the latent at level $i$ is
the negative log-probability of $\mb{y}_i$,
that is $\mc{L}_{\text{flow}} (\mb{y}_i) = -\log p_Y({\mb{y}_i})$, using
the change of variables as
\begin{equation}
    \log p_{Y}(\mb{y}_i)
    = \log p_{Z}(\mb{z}_i) + \log \left\vert \det \nabla_Y
g_i(\mb{y}_i) \right\vert
    = \log p_{Z}(\mb{z}_i) + \log \left\vert J_{Y} \left(
g_i(\mb{y}_i) \right) \right\vert\ ,
\end{equation}
where
\begin{equation}
    \log p_Z(\mb{z}_i) = -\frac{1}{d_i} \log
 \frac{1}{\left(\sqrt{2 \pi}\right)^{d_i}} \exp \left( -\frac{1}{2}
    \| \mb{z}_i \|^2 \right)
= \frac{1}{2} \log{2 \pi} + \frac{1}{2 d_i}
\| \mb{z}_i \|^2\ ,
\end{equation}
$d_i$ is the dimension of the $i$th latent and
$J_X(\cdot)$ computes the Jacobian matrix over
the partial derivative $X$.
Similarly, the log-probability of $\mb{x}_i$ at level $i$ is
\begin{equation}
\begin{aligned}
    \log p_{X} (\mb{x}_i)
    &= \log p_{Z}(\mb{z}_i)
    + \log \left\vert \det \nabla_X \left( g_i \circ
f_i (\mb{x}_i) \right) \right\vert \\
    &= \log p_{Z}(\mb{z}_i) + \log \left\vert \det J_{Y}
(g_i(\mb{y}_i)) \right\vert + \log \left\vert \det J_{X}
(f_i(\mb{x}_i)) \right\vert\ .
\end{aligned}
\end{equation}
Then, the log-probability of an image at level $i$ with hierarchical
autoencoders and flows from multiple downsampling layers,
$\mathbf{x}_{i+1} = d(\mathbf{x}_i)$ at level $i$, can be calculated
with the chain rule as
\begin{equation}
    \log p(\mathbf{x}_i) = \sum^i_{j=1} \log p_{X} (\mathbf{x}_j) + \log
\left\vert \det J_X(d(\mathbf{x}_{j-1})) \right\vert \cdot \mathbf{1} \left[ j>1 \right]\ ,
\end{equation}
where $[\cdot]$ is a binary indicator.

\subsection{Diffusion Models}
\revision{
\label{sec:dm}
Different from normalising flow models that sample in a low-dimensional
embedding space due to the otherwise large computational complexity, diffusion
models diffuse every image pixel in the image space independently,
enabling pixelwise sampling from the Gaussian distribution.
We outline below a strategy and formulas to allow uneven or extended
step diffusion in the backward diffusion process.

\subsubsection{Multi-step Diffusion Sampling}
\noindent \textbf{Forward process.}
The standard description of denoising diffusion model \citep{ddpm} defines a sequence of 
random variables
$\{ \v x_0, \v x_1, \ldots, \v x_T \}$ according to a forward diffusion process
\begin{equation}
\label{eq:forward recurrence}
\v x_{t+1} = \sqrt{\alpha_t} \, \v x_{t} + \sqrt{\beta_t}\, \epsilon\ ,
\end{equation}
where $\beta_t = 1 - \alpha_t$, $\v x_t$ is a sample from a 
random variable $X_t$, and
$\epsilon$ is a sample from the standard (multidimensional) Gaussian.
The index $t$ takes integer values between $0$ and $T$, and
the set of random variables form a Markov chain.

The idea can be extended to define a continuous family of random variables
according to the rule
\begin{equation}
\label{eq:continuous-recurrence}
\v x_t = \sqrt{\bar\alpha_t} \, \v x_0 + \sqrt{\bar\beta_t}\, \epsilon\ ,
\end{equation}
where $\bar\beta_t = 1 - \bar\alpha_t$, and for simplicity, we can assume that $x_t$ is defined for $t$ taking continuous
values in the interval $[0, 1]$.
Here, the values $\bar\alpha_t$ are a decreasing function of 
$t$ with $\bar\alpha_0 = 1$
and $\bar\alpha_1 = 0$. It is convenient to refer to $t$ as {\em time}.

It is easily seen that if $\{ 0 = t_0, t_1, ... , t_T = 1 \}$ are an increasing set of 
time instants between $0$ and $1$, then the sequence of random variables
$\{ X_{t_0}, \ldots X_{t_T} \}$ form a Markov chain.
Indeed, it can be computed that for $0 \le s < t \le 1$, the conditional
probabilities $p(\v x_t | \v x_s)$ are Gaussian  
\begin{equation}
p(\v x_t | \v x_s) = \calN \big(\v x_t ~|~ \sqrt{\bar\alpha_{st}} \v x_s, ~
\bar\beta_{st}\big) ~.
\end{equation}
where $\bar\alpha_{st} = \bar\alpha_t/\bar\alpha_s$ and $\bar\beta_{st}= 1-\bar\alpha_{st}$.  This is the 
isotropic normal distribution
having mean $\sqrt{\bar\alpha_{st}} \v x_s$ and variance 
$\bar\beta_{st}$.
Similarly to Eq.~\eqref{eq:forward recurrence}, one has
\begin{equation}
\label{eq:continuous-recurrence-accumulated}
\v x_t = \sqrt{\bar\alpha_{st}} \, \v x_s + \sqrt{\bar\beta_{st}}\,  \epsilon\ .
\end{equation}
This applies in particular when $s$ and $t$ refer to consecutive time
instants $t_i$ and $t_{i+1}$.
In this case, the joint probability of $\{ X_{t_0}, \ldots, X_{t_T} \}$ is given by
\begin{equation}
p(\v x_{t_0}, \v x_{t_1}, \ldots, \v x_{t_T}) = 
p(\v x_{t_0}) ~ \prod_{i=1}^T ~ p(\v x_{t_i} | \v x_{t_{i-1}}) ~.
\end{equation}
One also observes, from Eq.~\eqref{eq:continuous-recurrence} that $p(\v x_1)$ is 
a standard Gaussian distribution.
A special case is where the time steps are chosen evenly spaced between
$0$ and $1$. Thus, if $h = 1 / T$,  this can be written as
\begin{equation}
p(\v x_0, \v x_h, \v x_{2h}, \ldots, \v x_{Th}) = 
p(\v x_0) ~ \prod_{i=1}^T ~ p(\v x_{ih} | \v x_{(i-1)h}) ~.
\end{equation}
\noindent \textbf{Backward process.}
The joint probability distribution is also a Markov chain, which can
be written in the reverse order, as
\begin{equation}
p(\v x_{t_0}, \v x_{t_1}, \ldots, \v x_{t_T}) = 
p(\v x_{t_T}) ~ \prod_{i=1}^T ~ p(\v x_{t_{i-1}} | \v x_{t_{i}}) ~.
\end{equation}
This allows us to generate samples from $X_0$ by
choosing a sample from $X_1 = X_{t_T}$ (a standard Gaussian distribution)
and then successively sampling from the conditional probability
distributions $p(\v x_{t_{i-1}} | \v x_{t_{i}})$.

Unfortunately, although the forward conditional distributions 
$p(\v x_{t_i} | \v x_{t_{i-1}})$ are known Gaussian distributions,
the backward distributions are not known and are not Gaussian.
In general, for $s < t$, the conditional distribution $p(\v x_t | \v x_s)$ is
Gaussian, but the inverse $p(\v x_s | \v x_t)$ is not.

However, if $(t-s)$ is small, or more exactly, if the variance of the added noise,
$\bar\beta_{st} = 1 - \bar\alpha_{st}$
is small, then the distributions can be accurately approximated by Gaussians
with the same variance $\bar\beta_{st}$ as the forward conditionals.
With this assumption, the form of the backward conditional $p(\v x_s|\v x_t)$
is specified just by determining its mean, denoted by
$\mu(\v x_s | \v x_t)$.  The training process of the diffusion model consists
of learning (using a neural network) the function $\mu(\v x_s | \v x_t)$
as a function of $\v x_t$.  As explained in \citet{ddpm}, it is not necessary
to learn this function for all pairs $(s, t)$, as will be elaborated below.

We follow and generalize the formulation in \cite{ddpm}. The training
process learns a function $\epsilon_\theta(\v x_t, t)$ that minimizes
the expected least-squared loss function 
\begin{equation}
\Ex_{\v x_0 \sim X_0, \epsilon \sim \calN} ~
[\|\epsilon - \epsilon_\theta (\v x_t, t) \|^2]\ ,
\end{equation}
where $\v x_t = \sqrt{\bar\alpha_t} \v x_0 + \sqrt{\bar\beta_t} \epsilon$.  
As such
it estimates (exactly, if the optimum function $\epsilon_\theta$ is found) 
the expected value of the added noise, given $\v x_t$
(note that it estimates the {\em expected value} of the added noise,
and not the actual noise, which cannot be predicted).
In this case, following \cite{ddpm},
\begin{equation}
\mu(\v x_{t-1} | \v x_t) = \sqrt{\frac{\bar\alpha_{t-1}}{\bar\alpha_t}}
 \left( 
\v x_t - \frac{1 - \bar\alpha_{t}/\bar\alpha_{t-1}}{\sqrt{1 - \bar\alpha_t}}
\epsilon_\theta (\v x_t, t)
\right) ~.
\end{equation}
In this form, this formula is easily generalized to
\begin{equation}
\label{eq:backward-mean}
\mu(\v x_s | \v x_t) = \frac{1}{\sqrt{\bar\alpha_{st}}}
 \left( 
\v x_t - \frac{1-\bar\alpha_{st}}{\sqrt{1 - \bar\alpha_t}}
\epsilon_\theta (\v x_t, t)
\right)
= 
 \frac{1}{\sqrt{\bar\alpha_{st}}}
 \left( 
\v x_t - \frac{\bar\beta_{st}}{\sqrt{\bar\beta_t}}
 \epsilon_\theta (\v x_t, t)
\right) ~.
\end{equation}
As for the variance of $p(\v x_s | \v x_t)$, in \cite{ddpm}
it is assumed that the $p(\v x_{t-1} | \v x_t)$ is an isotropic Gaussian (although
in reality, it is not exactly a Gaussian, nor exactly isotropic).  The covariance matrix of
this Gaussian is denoted
by $\sigma_{st}^2 \mb I$, and two possible choices are given, which are generalized naturally
to 
\begin{equation}
\label{eq:backward-covariance}
\sigma^2_{st} = \bar\beta_{st} \text{~~~~~or~~~~~}
\sigma^2_{st} = \frac{\bar\beta_s \bar\beta_{st}}{\bar\beta_t} ~.
\end{equation}
As pointed out in \cite{ddpm} both of these are compromises.  The first choice
expresses the approximation that the variance of the noise added in the 
backward process is equal to the variance in the backward process.  As mentioned,
this is true for small time steps.  

Thus, in our work, we choose to model the reverse conditional as follows,
\begin{equation}
p_\theta(\v x_s | \v x_t ) = \calN (\v x_s | \mu (\v x_s | \v x_t),  \sigma^2_{st} \mb I)\ ,
\end{equation}
where $\mu(\v x_s | \v x_t)$ is given by Eq.~\eqref{eq:backward-mean} and
$\sigma^2_{st}$ is given by Eq.~\eqref{eq:backward-covariance}.  This is
an approximation of the true conditional probability $p(\v x_s | \v x_t)$.
} 

\subsubsection{Probability Estimation}
\revision{
In the following, we choose a finite set of $T$ time instances (usually
equally spaced)
$\{ 0 = \tau_0, \tau_1, \ldots, \tau_T = 1 \}$ and 
consider the Markov chain consisting of the variables $X_{\tau_t}$, for $t = \{0, \ldots, T\}$, at these
time instances.  For simplicity, we use the notation $X_t$ instead of $X_{\tau_t}$
and $\v x_t$ a sample from the corresponding random variable.
Then, the notation corresponds to the common notation in the literature,
but also applies in the case of unevenly, or widely sampled time instants. 

To distinguish between the true probabilities of the variables $X_t$
and the modelled conditional probabilities, the true probabilities will
be denoted by $q$ (instead of $p$ which was used previously). The modelled
probabilities will be denoted by $p_\theta(\v x_{t-1} | \v x_t)$, and
the probability distribution of $X_T$, which is Gaussian, will be denoted
by $p(\v x_T)$.
}

The image probability can be calculated by using the forward and
backward
processes for each step of a pretrained diffusion model. The joint probability
$p(\mb{x}_{0:T})$ and the probability of clean input $\mb{x}_0$ can be
computed using the forward and backward conditional probability,
$q(\mb{x}_{t+1} \vert \mb{x}_{t})$ and $p_{\theta}(\mb{x}_{t} \vert
\mb{x}_{t + 1})$ respectively.
Each sampling pair $(\v x_t, \v x_{t + 1})$ where $t \in S = \{0, 1, 2, ..., T - 1\}$, follows the Markov chain rule resulting in the joint probability
\begin{equation}
    p(\mb{x}_{0:T})
    = q(\mb{x}_0) \prod_{t \in S} q(\mb{x}_{t+1} \vert \mb{x}_{t})
    = p(\mb{x}_T) \prod_{t \in S} p_{\theta}(\mb{x}_{t} \vert \mb{x}_{t+1})\ ,
\end{equation}
so
\begin{equation}
    q(\mb{x}_0) = \frac{p(\mb{x}_T) \prod_{t \in S} p_{\theta}(\mb{x}_t
\vert
\mb{x}_{t + 1})}{\prod_{t \in S} q(\mb{x}_{t+1} \vert
\mb{x}_t)}\ .
\end{equation}

The negative log-probability of the input image $\mb{x}_0$ is then
\begin{equation}
\label{eq:diffusion_logpx}
-\log q(\mb{x}_0)
    = -\log p(\mb{x}_T) + \sum_{t \in S} \left(\underbrace{\log
q(\mb{x}_{t+1} \vert \mb{x}_t)}_{\text{forward process}} \underbrace{-\log
p_{\theta}(\mb{x}_t \vert \mb{x}_{t + 1})}_{\text{backward process}} \right)\ .
\end{equation}

\revision{
Computing Eq.~\eqref{eq:diffusion_logpx} can be decomposed into three steps:

\textit{1) Calculating $\log p(\mb{x}_T)$.}
Since $\mb{x}_0$ is fully diffused after $T$ forward steps,
$\mb{x}_T$ follows the standard Gaussian distribution $\mc{N}(\mb{0},
\mb{1})$,
the negative log-likelihood only depends on the Gaussian noises.

\textit{2) Calculating $\log q(\mb{x}_{t+1} \vert \mb{x}_t)$.}
Since $q(\v x_{t-1} | \v x_t)$ is a Gaussian with known mean $\bar\alpha_{t}/\bar\alpha_{t-1}$, and variance $1 - \bar\alpha_t/\bar\alpha_{t-1}$, 
the conditional probability is easily computed, as a Gaussian probability.

\textit{3) Calculating $\log p_{\theta}(\mb{x}_t \vert \mb{x}_{t + 1})$.}
Similarly, the probability $p_\theta(\v x_{t-1} | \v x_t)$ is
modelled as a Gaussian, with mean and variance given by
Eq.~\eqref{eq:backward-mean} and Eq.~\eqref{eq:backward-covariance} (where $s = t-1$)
the backward conditional probabilities are easily computed.
} 
 
\revision{\subsubsection{Higher-order Solution}
With the hypothesis that high-fidelity image generation is capable of
maintaining image semantics, in each of the diffusion inversion steps the
$\mb{x}_0$ estimation and log-likelihood calculation should be stable and
reliable with a small distribution variance.
The diffusion inversion, however, usually requires a sufficiently small
sampling
step $h$, where DDPM~\citep{ddpm} only supports $h=1$ and DDIM
is vulnerable to $h$~\citep{ddim} as evidenced in
Fig.~\ref{fig:ddim_rk4withddim}.
It is important to alleviate the effect of $h$ on the generation step
by stabilizing the backward process in diffusion models.

Without loss of generality, the Runge-Kutta method (RK4)~\citep{RK41,RK42}
can achieve a stable inversion process by constructing a higher-order function
to solve an initial value problem.
Different from the traditional RK4, the diffusion inversion requires
inverse-temporal updates because of the denoising gradient direction
from the initial noisy image at $t=T$ to the clean image at $t=0$.
We provide the formulation of traditional RK4 and our inverse-temporal
version in the appendix.}

\subsection{Experiments}
\revision{For each of the hierarchical normalising flows (NFs) and
diffusion models (DMs), we first show the effectiveness of
likelihood estimation to analyse the image distribution (on 2D
images for NFs and point samples for DMs).
For likelihood estimation with image fidelity, we then illustrate the
quality of images generated by our generation models (sampling on the
manifold from a Gaussian distribution as well as resolution enhancement
in NFs and sampling step exploration with RK4 stabilisation in DMs).}

\subsubsection{Experiments on Hierarchical Normalising Flow Models}

\textbf{Probability Estimation.}
Fig.~\ref{fig:celeba_pae} illustrates the probability density estimation on
level 3 for an in-distribution dataset CelebA~\citep{liu2015faceattributes} and an out-of-distribution
dataset CIFAR10~\citep{cifar10}.
The distribution of the latent variable $\mb{z}_i$ of CelebA is concentrated
on
a \revision{higher} mean value than that of CIFAR10 due to the learning
of $\mb{z}_i$
in
the standard Gaussian distribution.
Similarly, this distribution tendency is not changed in the image space
illustrated by $\log p(\mb{x}_i)$.
In this case, outlier samples from the in-distribution dataset can be detected
with a small probability in the probability estimation.
\setcounter{figure}{1}
\begin{subfigure}
\centering
\setcounter{subfigure}{0}
\begin{minipage}[b]{0.32\textwidth}
    \includegraphics[width=\linewidth]{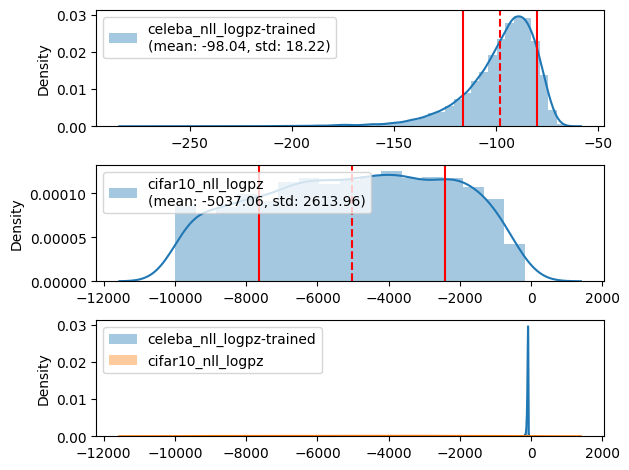}
    \itsubcaption{$\log p(\mb{z})$}
\end{minipage}
\begin{minipage}[b]{0.32\textwidth}
    \includegraphics[width=\linewidth]{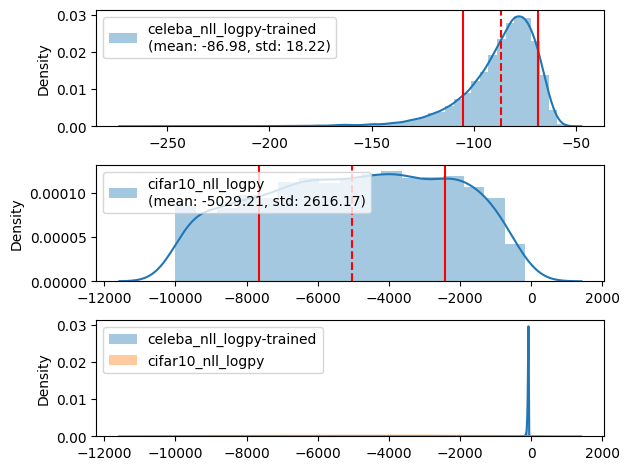}
    \itsubcaption{$\log p(\mb{y})$}
\end{minipage}
\begin{minipage}[b]{0.32\textwidth}
    \includegraphics[width=\linewidth]{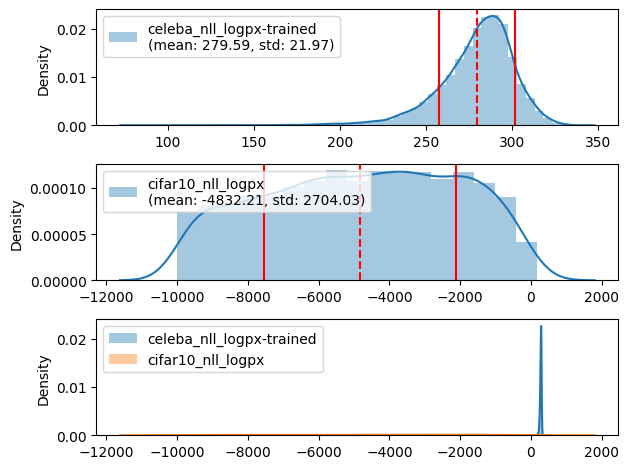}
    \itsubcaption{$\log p(\mb{x})$}
\end{minipage}
\setcounter{subfigure}{-1}
\itcaption{Log-likelihood estimation using hierarchical autoencoders and
flows.
The encoder and flow are trained on CelebA and evaluated on CelebA
and CIFAR10.
The x-axis is $\log p(\cdot)$ and the y-axis is the histogram density.
In each subfigure, the first row is on the in-distribution dataset CelebA and
the
second row is on out-of-distribution CIFAR10, both are in the last row.
In \textbf{(A)}, $\log p(\mb{z})$ can detect outlier samples, and adding
$\log \vert
\det (\cdot) \vert$ from NF and autoencoder does not significantly
affect
the distribution tendency, see \textbf{(B)} and \textbf{(C)}.
For better visualisation, samples with $\log p(\cdot)$ less than -10,000 are
filtered out.}
\label{fig:celeba_pae}
\end{subfigure}

\textbf{Random Image Generation.}
Image reconstructions with encoded latent variables and conditional images as
well as random samples are provided in Fig.~\ref{fig:decoding_joint_celeba}.
For the low-level autoencoder and flow, say at level 1, conditioned on the
sequence of decoded $\mb{x}_i$ for $i=\{2,3,4\}$, the reconstruction of
$\mb{x}_1$ is close to the
processed images although some human facial details are lost due to the
downsampling mechanism, see
Fig.~\ref{fig:decoding_joint_celeba}\textbf{(A)}.
While randomly sampling $\{\mb{z}_i\}$ from the normal distribution at
each
level, the generated
human faces are smooth but with blurry details in such as hair and chin and
lack
a realistic background.

\textbf{Image Super-resolution.}
\setcounter{figure}{3}
\begin{subfigure}
\centering
\setcounter{subfigure}{0}
\begin{minipage}[b]{0.48\textwidth}
    \centering
    \includegraphics[width=\linewidth]{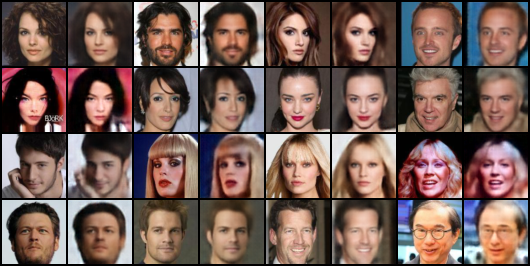}
    \itsubcaption{Reconstruction \revision{at level 1} with $\{z_i\}$
    from encoders $\{g_i\}$
and \revision{conditioned on} $\{f'_i\}$.}
\end{minipage}
\begin{minipage}[b]{0.48\textwidth}
    \centering
    \includegraphics[width=\linewidth]{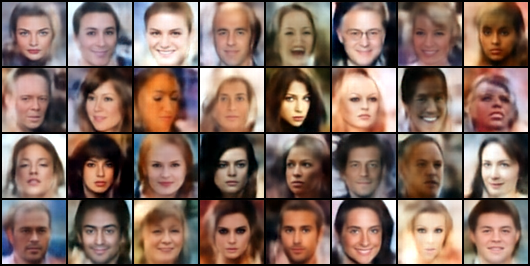}
    \itsubcaption{Random generation \revision{at level 1} with latent variables
$\{z_i\}\sim\mathcal{N}(\mb{0},\mb{1})$ and \revision{conditioned on}
$\{f'_i\}$.}
\end{minipage}
\setcounter{subfigure}{-1}
\itcaption{Image \revision{reconstruction and} generation on the end-to-end
training of 4-level autoencoders
and flows. \revision{For each of two columns from left to right
in \textbf{(A)}, the left is the real image and the right is the
reconstructed image.}}
\label{fig:decoding_joint_celeba}
\end{subfigure}
With the jointly trained autoencoders and flows on CelebA,
the images with low resolution, $3\times8\times8$
(channel$\times$height$\times$width) and $3\times16\times16$, are decoded to
$3\times64\times64$ with smooth human faces, see
Fig.~\ref{fig:super_res_celeba_joint}\textbf{(A)} and
Fig.~\ref{fig:super_res_celeba_joint}\textbf{(B)} respectively.
The low-resolution image $\mb{x}_i$ is used as a condition image for 1) NF
inverse $\{g'_i\}$ to generate embedding code to combine with the randomly
sampled
$\mb{z}_i \sim \mathcal{N}(\mb{0}, \mb{1})$ and 2) decoders
$\{f'_i\}$ to concatenate with all upsampling layers in each decoder.
This preserves the human facial details from either high levels or low levels
for realistic image generation.
As the resolution of the low-resolution images increases, the embedding code
contains richer details.
\setcounter{figure}{4}
\begin{subfigure}[!ht]
\centering
\begin{minipage}[b]{0.48\textwidth}
    \centering
    \includegraphics[width=\textwidth]{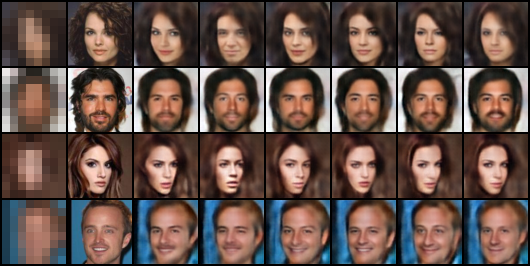}
    \itsubcaption{Resolution: $3\times8\times8$ to $3\times64\times64$}
\end{minipage}
\begin{minipage}[b]{0.48\textwidth}
    \centering
    \includegraphics[width=\textwidth]{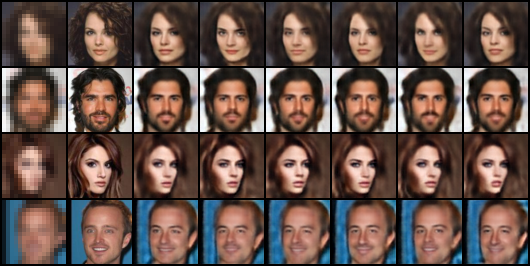}
    \itsubcaption{Resolution: $3\times16\times16$ to $3\times64\times64$}
\end{minipage}
\setcounter{subfigure}{-1}
\itcaption{Image super-resolution on dataset CelebA.
The first column is low-resolution images, the second column is real images,
and the rest are high-resolution images with latent variables
$\{Z_i\}\sim\mathcal{N}(\mb{0}, \mb{1})$ conditioned on the
low-resolution images and temperature $1.0$.}
\label{fig:super_res_celeba_joint}
\end{subfigure}

\subsubsection{Experiments on Diffusion Models}

\textbf{Log-likelihood Estimation on Point Samples.}
\begin{figure}[!ht]
\centering
\includegraphics[width=0.95\linewidth]{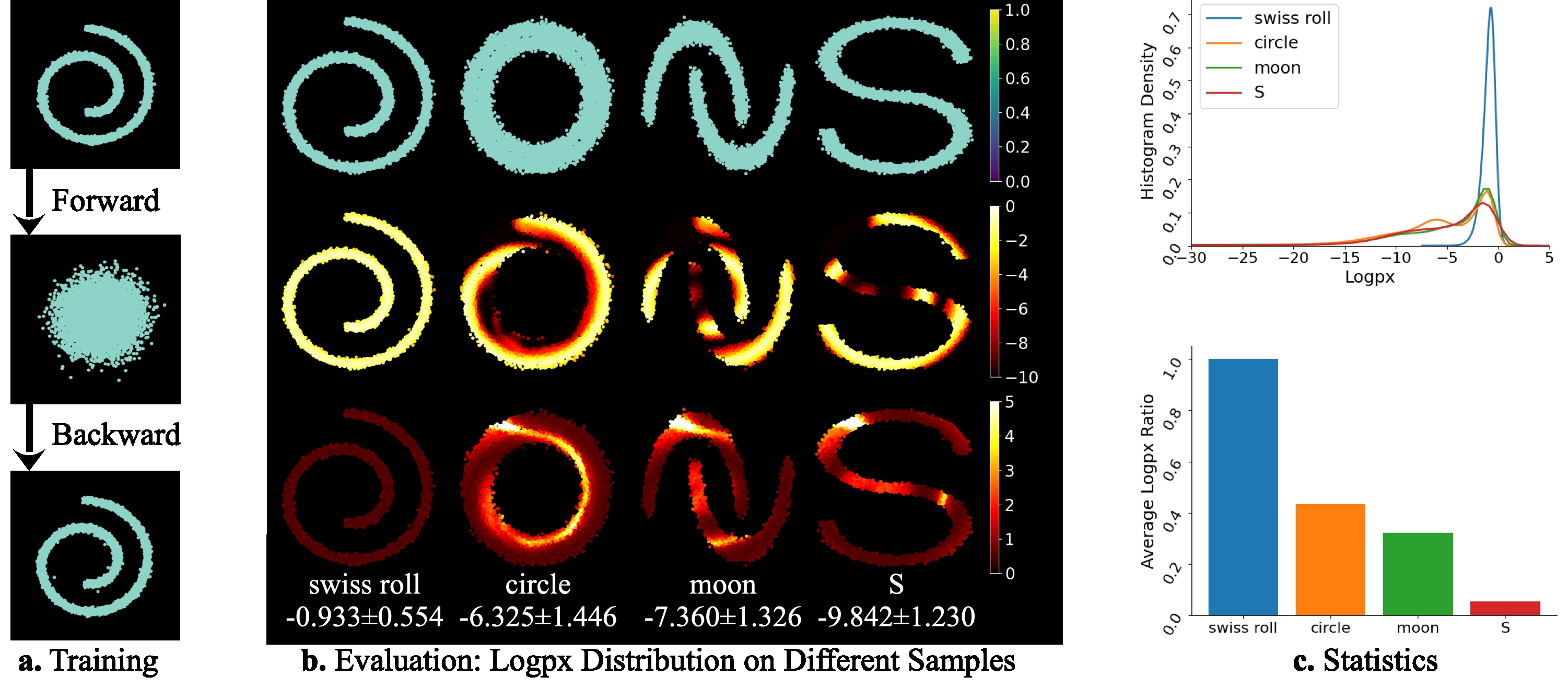}
\itcaption{Evaluation of log-probability of $\mb{x}_0$ on point samples with
each
of 10,000 points.
\textbf{(A)} The training is on a Swiss roll sample and a diffusion model with
forward (noising) and backward (denoising) processes.
\textbf{(B)} At the evaluation phase, unseen samples, that is circle, moon,
and
S, have lower $\log p(\mb{x})$ values than the seen Swiss roll sample.
In b, the first row is sampled points and the middle and last rows are the
mean
value and the standard deviation of $\log p(\mb{x})$ for each point on 100
random rounds respectively, which is represented as ``mean $\pm$ std".
The randomness lies in the random noise in the forward and backward
processes.
A lighter colour indicates a higher density.
\textbf{(C)} Statistics indicates the higher density of a seen sample (Swiss
roll) than an unseen one (circle, moon, or S) through the diffusion model by
using the negative of Eq.~\eqref{eq:diffusion_logpx} with $\log_{10}$.}
\label{fig:swiss_roll}
\end{figure}
We evaluate the log-probability of each point of point samples~\citep{scikit-learn} including Swiss
roll, circle, moon, and S shown in Fig.~\ref{fig:swiss_roll}.
Given a pretrained diffusion model on Swiss roll samples with 100 forward
steps
with each diffused by random Gaussian noise (see
Fig.~\ref{fig:swiss_roll}\textbf{(A)}, the log-probability of the samples in
Fig.~\ref{fig:swiss_roll}\textbf{(B)} follows Eq.~\eqref{eq:diffusion_logpx}
with $h=1$ and indicates higher probability and density on seen or
similar samples than unseen ones.
In Figs.~\ref{fig:swiss_roll}\textbf{(B)}-\textbf{(C)}, the mean value of the
Swiss roll sample achieves a higher mean value, -0.933, and a higher histogram
density, 0.7, than the others.
As the difference in the sample shape from the Swiss roll increases, the
log-likelihood decreases, as shown in the bar chart in
Fig.~\ref{fig:swiss_roll}\textbf{(C)}.
It indicates that sampling from a low-density distribution is unable to
reverse
the diffusion step to obtain a realistic sample from the training set.

\textbf{DDPM Sampling with Large Steps.}
\setcounter{figure}{5}
\begin{subfigure}
\centering
\setcounter{subfigure}{0}
\begin{minipage}[b]{0.24\textwidth}
    \includegraphics[width=\linewidth]{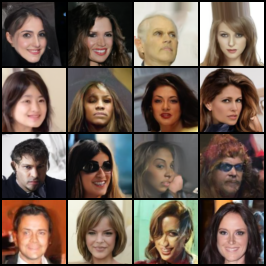}
    \itsubcaption{$h=1$}
\end{minipage}
\begin{minipage}[b]{0.24\textwidth}
    \includegraphics[width=\linewidth]{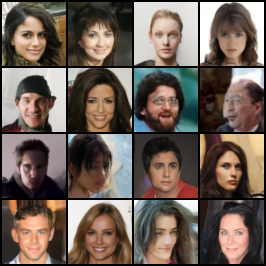}
    \itsubcaption{$h=2$}
\end{minipage}
\begin{minipage}[b]{0.24\textwidth}
    \includegraphics[width=\linewidth]{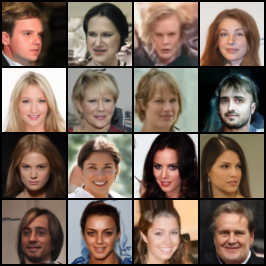}
    \itsubcaption{$h=10$}
\end{minipage}
\begin{minipage}[b]{0.24\textwidth}
    \includegraphics[width=\linewidth]{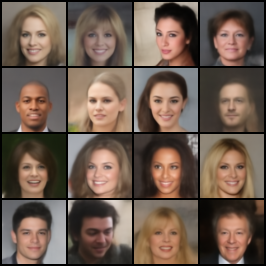}
    \itsubcaption{$h=100$}
\end{minipage}
\setcounter{subfigure}{-1}
\itcaption{Image generation from our modified DDPM with step size $h$.
Samples follow a Gaussian distribution.
Fine details are obtained even for very large steps ($h = 100$).
}
\label{fig:ddpm_steps}
\end{subfigure}
While Fig.~\ref{fig:swiss_roll} uses $h = 1$ as the standard DDPM
sampling process, it is feasible to sample with a fairly large step without
losing the sample quality.
This enables sampling from the Gaussian distribution for the log-likelihood
estimation with less running time.
To visualise the image quality, we evaluate the samples on CelebA dataset by
using a pretrained diffusion model with 1,000 forward diffusion steps.
In Fig.~\ref{fig:ddpm_steps}, the sampling has
an
increase step $h$ in \{2, 10, 100\} while the samples have a high quality for
$h=\{2, 10\}$ and a fair quality for $h=100$.

\begin{figure}[!ht]
\centering
\begin{minipage}[c]{0.6\textwidth}
    \includegraphics[width=\linewidth]{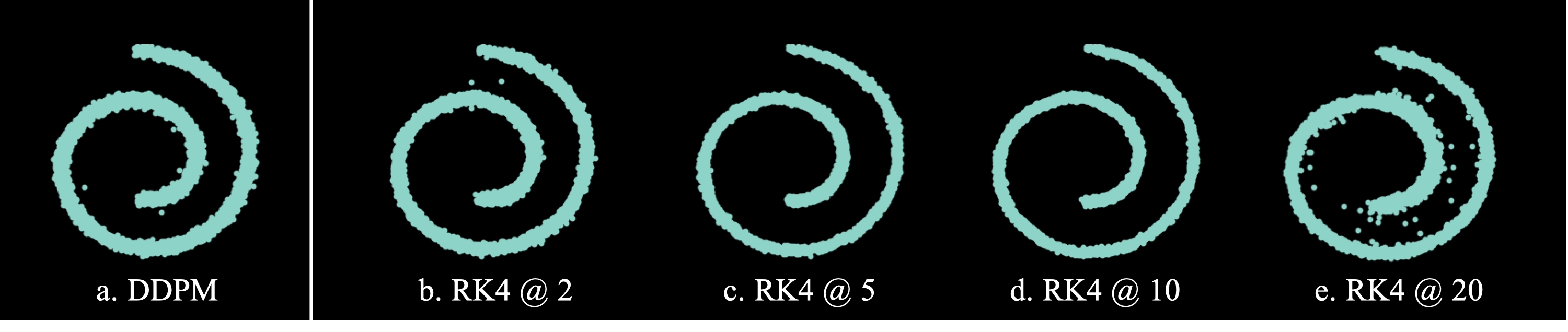}
\end{minipage}
\hfill
\begin{minipage}[c]{0.37\textwidth}
\itcaption{Sampling robustness of DDPM and RK4 @ step $h$.
With $h$ being 5 or 10, RK4 still achieves clear sampling compared with
DDPM.
If $h$ is too large, for instance 20, RK4 fails as expected.}
\label{fig:swiss_roll_rk4}
\end{minipage}
\end{figure}

\textbf{Higher-order Solution Stabilises Sampling.}
While sampling with a large step $h$ can sometimes cause bias from the
one with a small $h$, RK4 effectively alleviates such a bias.
We evaluate both the point samples and human face images from CelebA.
In Fig.~\ref{fig:swiss_roll_rk4}, compared with the sample by using DDPM, RK4
with DDPM inference achieves less noise at $h=\{2, 5, 10\}$.
For $h=20$, RK4 performs expectedly worse because it only applies 5
sampling steps while the training is on $(T=100)$ diffusion steps.
In Fig.~\ref{fig:ddim_rk4withddim}, we apply DDIM as the inference method for
RK4 to deterministically compare the samples with DDIM.
As $h$ increases from 1 to 100, many of the samples using DDIM lose the
image consistency with the samples at $h=1$;
however, most of the samples using RK4 still retain the image consistency.
This indicates the robustness of applying RK4 with a large sampling step.

\begin{subfigure}[!ht]
\setcounter{figure}{8}
\centering
\setcounter{subfigure}{0}
\begin{minipage}[b]{0.24\textwidth}
    \includegraphics[width=\linewidth]{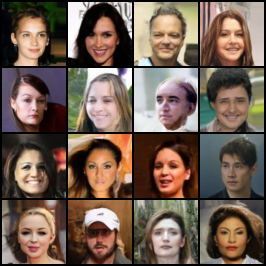}
    \itsubcaption{DDIM@1}
\end{minipage}
\begin{minipage}[b]{0.24\textwidth}
    \includegraphics[width=\linewidth]{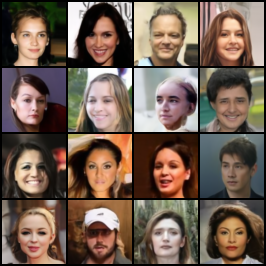}
    \itsubcaption{DDIM@2}
\end{minipage}
\begin{minipage}[b]{0.24\textwidth}
    \includegraphics[width=\linewidth]{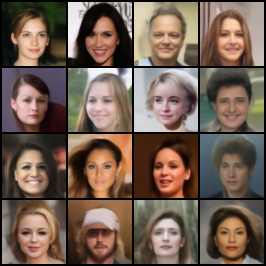}
    \itsubcaption{DDIM@10}
\end{minipage}
\begin{minipage}[b]{0.24\textwidth}
    \includegraphics[width=\linewidth]{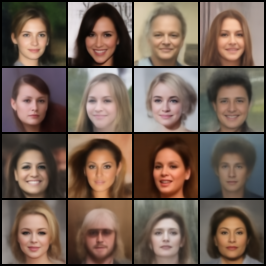}
    \itsubcaption{DDIM@100}
\end{minipage}
\\
\begin{minipage}[b]{0.24\textwidth}
    \includegraphics[width=\linewidth]{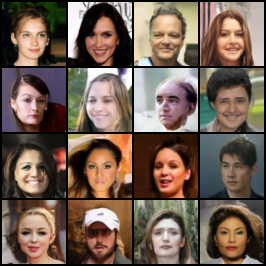}
    \itsubcaption{RK4@1}
\end{minipage}
\begin{minipage}[b]{0.24\textwidth}
    \includegraphics[width=\linewidth]{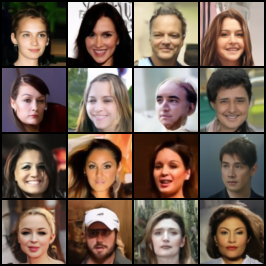}
    \itsubcaption{RK4@2}
\end{minipage}
\begin{minipage}[b]{0.24\textwidth}
    \includegraphics[width=\linewidth]{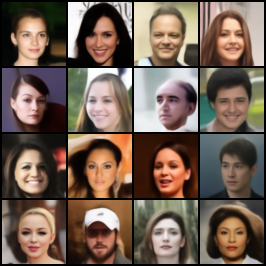}
    \itsubcaption{RK4@10}
\end{minipage}
\begin{minipage}[b]{0.24\textwidth}
    \includegraphics[width=\linewidth]{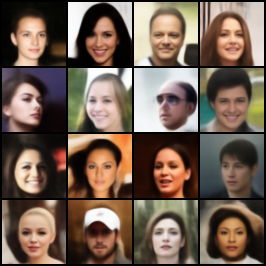}
    \itsubcaption{RK4@100}
\end{minipage}
\setcounter{subfigure}{-1}
\itcaption{Random image generation using DDIM and RK4 with DDIM as inference @
time step $h$=\{1, 2, 10, 100\}.
The RK4 sampling method is more robust than DDIM, especially at $h=100$,
with a higher image consistency than those at $h=1$.
}
\label{fig:ddim_rk4withddim}
\end{subfigure}

%
\section{Semantic Disentanglement on Manifold}
\label{sec:el}
\revision{Semantics of object attributes are crucial for image distribution
and spatial presentation.
For instance, different shapes in Fig.~\ref{fig:swiss_roll} represent different
objects while those closer to the seen samples have high likelihood; in
Fig.~\ref{fig:ddim_rk4withddim} semantics such as human gender (see the 2nd
row and 3rd column image with DDIM and RK4) are fundamental for controllable
generation by sampling in high-density regions of specific semantic clusters
on the manifold.
These semantics, however, are usually entangled without independent
distributions from each other for deterministic embedding sampling on the image manifold~\citep{disentanglement_vae_1,disentanglement_vae_2,disentanglement_vae_3}.
Hence, regardless of image generation models, we exploit the popular and
efficient variational autoencoder and introduce our GridVAE model for effective
semantic disentanglement on the image manifold.}

\subsection{GridVAE for Clustering and Disentanglement}
\label{sec:gridvae_disen}
\subsubsection{Formulation}
A variational autoencoder (VAE)~\citep{kingma2013auto} is a neural network
that
maps inputs to a distribution instead of a fixed vector. Given an input $\mb{x}$,
the
encoder with neural network parameters $\phi$ maps it to a hidden
representation
$\mb{z}$. The decoder with the latent representation $\mb{z}$ as its input and the
neural
network parameters as $\theta$ reconstructs the output to be as similar to the
input $\mb{x}$. We denote the encoder $q_\phi (\mb{z}|\mb{x})$ and decoder $p_\theta (\mb{x}|\mb{z})$.
The hidden representation follows a prior distribution $p(\mb{z})$.

With the goal of making the posterior $q_\phi (\mb{z}|\mb{x})$ close to the actual
distribution $p_\theta (\mb{z}|\mb{x})$, we minimise the Kullback-Leibler divergence
between these two distributions. Specifically, we aim to maximise the
log-likelihood of generating real data while minimising the difference between
the real and estimated posterior distribution by using the evidence lower
bound
(ELBO) as the VAE loss function
\begin{equation}
\label{eqn:vae}
L(\theta, \phi) = -\log p_\theta(\mb{x}) + D_{KL}(q_\phi(\mb{z}|\mb{x})||p_\theta(\mb{z}|\mb{x})) =
-\mathbb{E}_{\mb{z}\sim q_\phi(\mb{z}|\mb{x})} \log p_\theta(\mb{x}|\mb{z}) + D_{KL}(q_\phi(\mb{z}|\mb{x}) \|
p_\theta(\mb{z}))\ ,
\end{equation}
where
the first term is the reconstruction loss and the second term is the
regularisation for $q_\phi(\mb{z}|\mb{x})$ to be close to $p_\theta(\mb{z})$. The prior
distribution of $\mb{z}$ is often chosen to be a standard unit isotropic Gaussian,
which implies that the components of $\mb{z}$ should be uncorrelated and hence
disentangled. If each variable in the latent space is only representative of a
single element, we assume that this representation is disentangled and can be
well interpreted.

Emergent language (EL)~\citep{havrylov2017emergence} is hereby introduced as a
language that arises spontaneously in a multi-agent system without any
pre-defined vocabulary or grammar. EL has been studied in the context of
artificial intelligence and cognitive science to understand how language can
emerge from interactions between agents. EL has the potential to be
compositional such that it allows for referring to novel composite concepts by
combining individual representations for their components according to
systematic rules. However, for EL to be compositional, the latent space needs
to
be disentangled~\citep{chaabouni2020compositionality}. Hence, we integrate VAE
into the EL framework by replacing the sender LSTM with the encoder of the VAE
noting that the default LSTM encoder will entangle the symbols due to its
sequential structure where the previous output is given as the input to the
next
symbol. In contrast, the symbols can be disentangled with a VAE encoder.

To achieve disentangled representations in EL, the VAE encoder must be able to
cluster similar concepts into discrete symbols that are capable of
representing
attributes or concepts. The standard VAEs are powerful, but their prior
distribution, which is typically the standard Gaussian, is inferior in
clustering tasks, particularly the location and the number of cluster centres.
In the EL setting, we desire a posterior distribution with multiple clusters,
which naturally leads to an MoG prior distribution with $K$ components
\begin{equation}
    p(\mb{z})= \frac{1}{K}\sum_{k=1}^{K} \mathcal{N}(\mb{z}|\mu_k,\sigma_k^2)\ .
\end{equation}
We choose the $\mu_k$ to be located on a grid in a Cartesian coordinate system
so that the posterior distribution clusters can be easily determined based on
the sample's distance to a cluster centre. We refer to this new formulation as
GridVAE, which is a VAE with a predefined MoG prior on a grid. The
KL-divergence
term in Eq.~\eqref{eqn:vae} can be re-written as
\begin{equation}
    D_{KL}( q_\phi(\mb{z}|\mb{x}) \| p_\theta(\mb{z})) = \mathbb{E}_{\mb{x}\sim
p(\mb{x})}\mathbb{E}_{q_\phi(\mb{z}|\mb{x})} [\log p(\mb{z}) - \log q_\phi(\mb{z}|\mb{x})]\ .
\end{equation}

The log probability of the prior can be easily calculated with the MoG
distribution, and we only need to estimate the log probability of the
posterior
using a large batch size during training. By using a GridVAE, we can obtain a
posterior distribution with multiple clusters that correspond to the same
discrete attribute, while allowing for variations within the same cluster to
generate different variations of the attribute.

\subsubsection{Experiments}
We evaluate the clustering and disentanglement capabilities of the proposed
GridVAE model using a two-digit MNIST dataset~\citep{mnist} consisting of
digits
0 to 5. Each digit is from the original MNIST dataset, resulting in a total of
36 classes [00, 01, 02, ..., 55].

To extract features for the encoder, we use a 4-layer
ResNet~\citep{he2016deep}
and its mirror as the decoder. The VAE latent space is 2-dimensional (2D), and
if the VAE learns a disentangled representation, each dimension of the 2D
latent
space should represent one of the digits. We use a 2D mixture of Gaussian
(MoG)
as the prior distribution, with 6 components in each dimension centred at
integer grid points from [-2, -1, 0, 1, 2, 3], that is the coordinates for the
cluster centres are [(-2, -2), (-2, -1), ..., (3, 3)]. The standard deviation
of
the mixture of Gaussian is $1/3$.

\begin{figure}[!ht]
\centering
\begin{minipage}{.3\linewidth}
    \centerline{\includegraphics[width=0.9\linewidth]{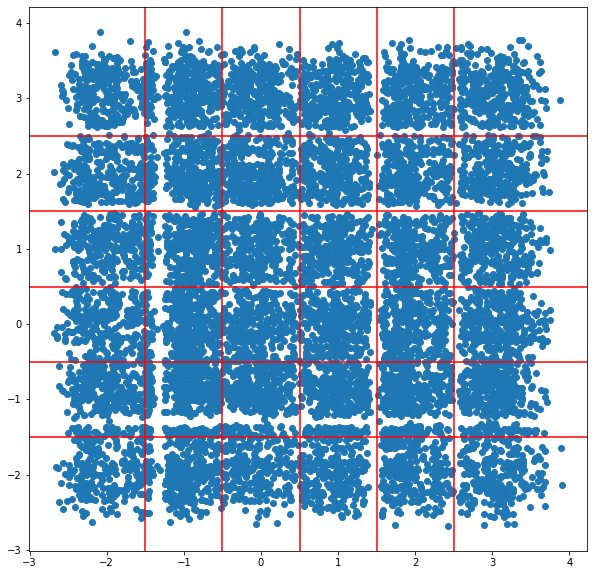}}
    \itcaption{Scatter plot of test set latent space with an MoG prior.}
    \vspace{16mm}
    \label{fig:clusters}
\end{minipage}
\hfill
\begin{minipage}{.68\linewidth}
    \centerline{\includegraphics[width=0.42 \linewidth]{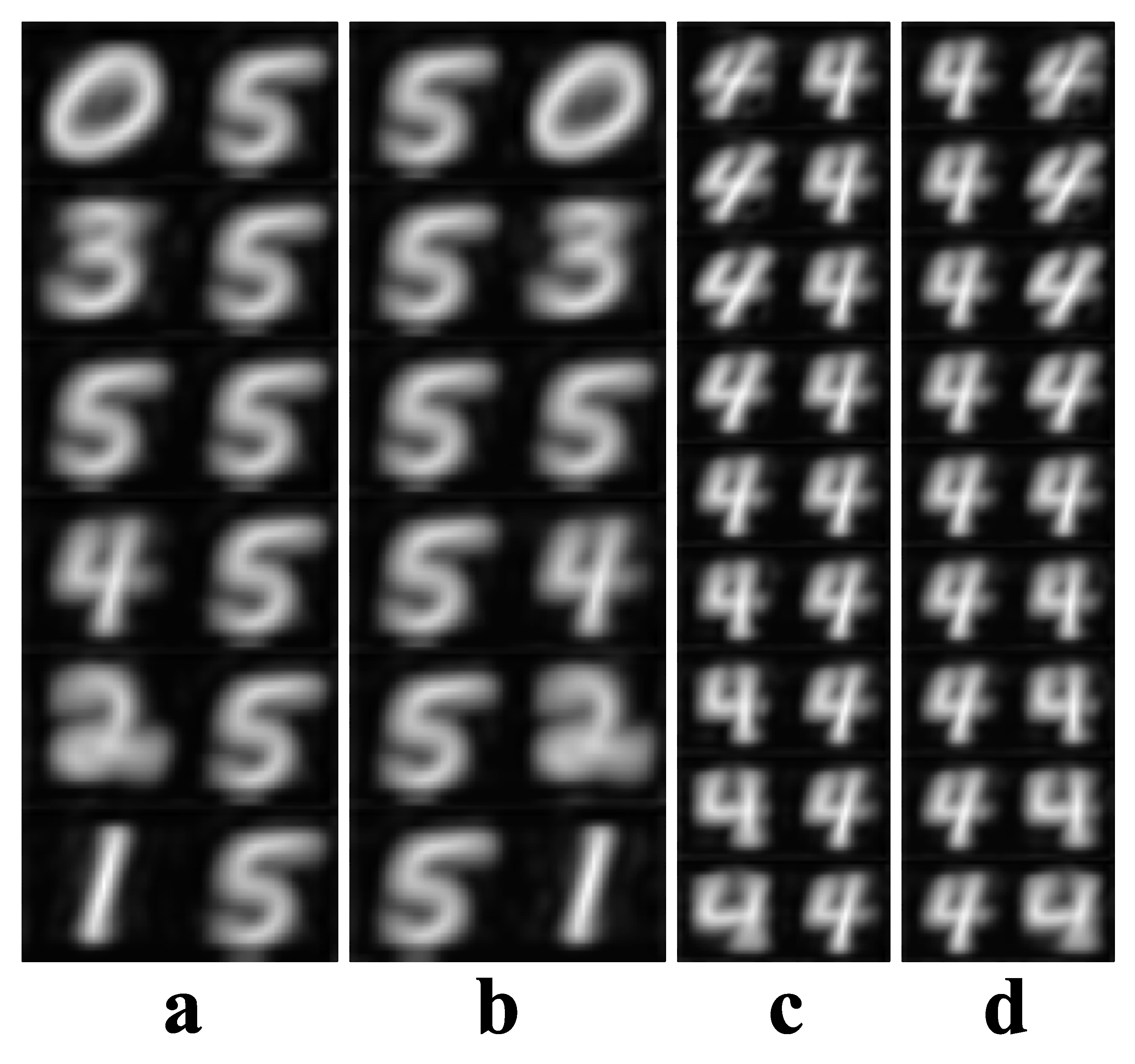}}
    \itcaption{Generated images from sampling the latent space. \textbf{(A)}
The
second dimension is fixed at 0, changing the first dimension from -2 to +3.
\textbf{(B)} The first dimension is fixed at 0 and the second dimension is
changed from -2 to +3. \textbf{(C)} Around the cluster centre(1, 1), keep the
second dimension fixed and change the first dimension. \textbf{(D)}
Around the
cluster centre(1, 1), keep the second dimension fixed and change the first
dimension.}
    \label{fig:decoder_plots}
\end{minipage}
\end{figure}

After training the model, we generate a scatter plot of the test set latent
space, as shown in Fig.~\ref{fig:clusters}. Since the prior is a mixture of
Gaussian on the grid points, if the posterior matches the prior, we can
simply
draw a boundary in the middle of two grid points, illustrated by the red
lines
in Fig.~\ref{fig:clusters}.

With the trained model, one can sample in the latent space for image
generation.
In Figs.~\ref{fig:decoder_plots} \textbf{(A)}-\textbf{(B)}, when we decode
from
the cluster centres $(i, j)$: in \textbf{(A)} we keep $j=0$ and change $i$
from
$-2$ to $3$, while in \textbf{(B)} we keep $i=0$ and change $j$ from $-2$ to
$3$. The latent space is disentangled with respect to the two digits - the
first
dimension of the latent space controls the first digit, while the second
dimension controls the second digit. Each of the cluster centres corresponds
to
a different number.

Figs.~\ref{fig:decoder_plots}\textbf{(C)}-\textbf{(D)} show images generated
within the cluster centred at $(1, 1)$, that is the pairs of number ``44". If
we
slightly modify one of the dimensions, it corresponds to different variations
of
the number ``4" along this dimension, while keeping the other digit unchanged.

Overall, these results demonstrate the effectiveness of the proposed GridVAE
model in clustering and disentangling the latent space on the two-digit MNIST
dataset.

\subsection{Scaling Up GridVAE}
In Sec.~\ref{sec:gridvae_disen}, the two-digit MNIST dataset lies in a
2-dimensional latent space. However, many real-world datasets would require a
much higher dimensional space.

\subsubsection{Addressing Higher Dimensional Latent Space}
Discretising a continuous space, such as in GridVAE, is challenging due to the
curse of dimensionality~\citep{curse_dim}. This refers to the exponential
growth
in the number of clusters as the number of dimensions increases, which leads
to
a computational challenge when dealing with high-dimensional latent space. For
example, when applying GridVAE to reconstruct images of the
CelebA~\citep{liu2015faceattributes} dataset to learn the 40 attributes, we
need
a 40-dimensional latent space with two clusters in each dimension to represent
the presence or absence of a given attribute. Firstly, parametrising the
mixture
of Gaussian prior $p(\mb{z})= \sum_{k=1}^{K} \mathcal{N}(\mb{z}|\mu_k,\sigma_k^2) / K$
over 40 dimensions is prohibitively expensive as $K=2^{40}\approx
1.1\times10^{12}$. Secondly, the assumption of equal probability for the
components, which was appropriate for the simple 2-digit MNIST dataset, is no
longer valid. This is because the attributes in the CelebA dataset are not
uniformly distributed, and some combinations may not exist. For instance, the
combination of ``black hair" + ``blonde hair" + ``brown hair" + ``bald" is
impossible due to attribute conflicts.
To address this issue, we use the proposed loss function in
Eq.~\eqref{eqn:vae}
incorporating relaxation.

\begin{figure}[!ht]
\centering
\begin{minipage}[c]{0.3\textwidth}
    \includegraphics[width=0.95\linewidth]{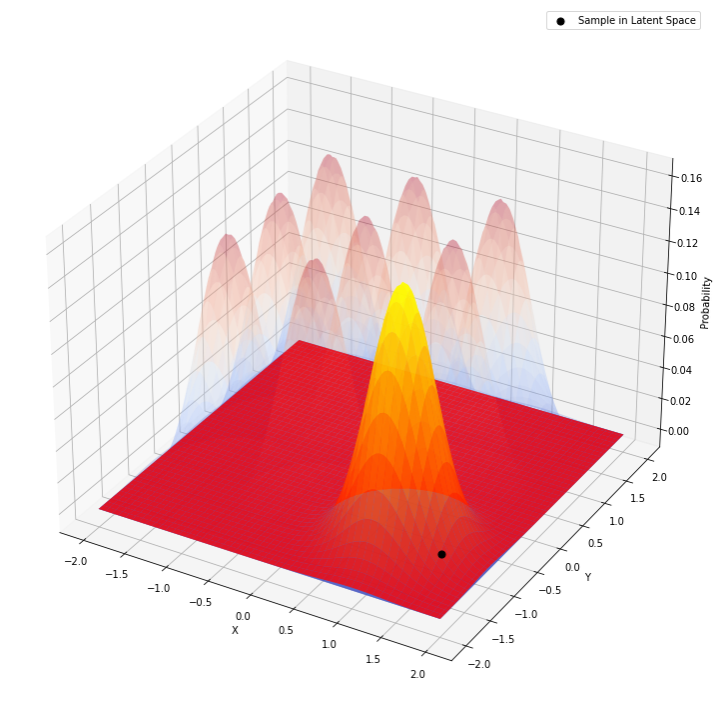}
\end{minipage}
\hfill
\begin{minipage}[c]{0.67\textwidth}
\itcaption{When calculating the KL-divergence, only the mixture component
closest to the data (darker shade) is considered. Other components (lighter
shade) are ignored. This can be generalised to multiple dimensions and
multiple
components in each dimension.}
\label{fig:new_prior}
\end{minipage}
\end{figure}
To avoid pre-parametrising $p(\mb{z})$ over 40 dimensions, we have implemented a
dynamic calculation of the KL-divergence between $q_\phi$ and $p_\theta$,
whereby only the cluster that is closest to the latent space representation is
considered, as illustrated in Fig.~\ref{fig:new_prior}. This means that
clusters
to which the data point does not belong do not affect its distribution,
and
the MoG distribution is simplified to a multivariate Gaussian as
\begin{equation}
D_\text{KL}(p_1\mid\mid p_2) = \frac{1}{2}\left[\log\frac{|\Sigma_2|}{|
\Sigma_1|} - n + \text{tr} {\left( \Sigma_2^{-1}\Sigma_1 \right)} + (\mu_2 -
\mu_1)^\intercal \Sigma_2^{-1}(\mu_2 - \mu_1)\right]\ ,
\end{equation}
where $p_1=q_\phi(\mb{z}|\mb{x})=\mathcal{N}(\mb{z}|\mu_1,\Sigma_1)$,
$\Sigma_1=\text{diag}(\sigma_1^2,\ldots,\sigma^2_n)$, $p_2 =
\mathcal{N}(\mu_2,\Sigma_2)$, $\mu_2=\texttt{R}(\mu_1)$, and
$\Sigma_2=\text{diag}(\sigma_0^2,\ldots,\sigma^0_n)$ with the round function
$\texttt{R}(\cdot)$ for the closest integer.

The key step here is that the round function dynamically selects the cluster
centre closest to $\mu_1$, and $\sigma_0$ is a pre-defined variance for the
prior distribution. It should be chosen so that two clusters next to each
other
have a reasonable degree of overlap, for example, $\sigma_0=1/16$ in some of
our
following experiments. The KL-divergence term becomes
\begin{equation}~\label{eq:vae_kl}
	\begin{aligned}
		D_{KL}( q_\phi(\mb{z}|\mb{x}) \| p_\theta(\mb{z}) )
		&=
		\frac{1}{2}\left[\log\frac{|\Sigma_2|}{|\Sigma_1|} - n + \text{tr} \left(
\Sigma_2^{-1}\Sigma_1 \right) + (\mu_2 - \mu_1)^\intercal \Sigma_2^{-1}(\mu_2 -
\mu_1)\right]\\
		&= \frac{1}{2}\left[\log\prod_i\sigma_0^2-\log\prod_i\sigma_i^2 - n +
\sum_i\frac{\sigma_i^2}{\sigma_0^2}  + \sum_i\frac{(\mu_i-\texttt{R}(
\mu_i))^2}{
\sigma_0^2} \right]\\
		&= \frac{1}{2}\left[\sum_{i=1}^{n}\left(\log\sigma_0^2 - \log\sigma_i^2 -
1\right) + \sum_{i=1}^n\frac{\sigma_i^2 +
(\mu_i-\texttt{R}(\mu_i))^2}{\sigma_0^2}  \right]\ .
	\end{aligned}
\end{equation}

By adopting Eq.~\eqref{eq:vae_kl}, we can significantly reduce the
computational
complexity of the model, even for a high-dimensional latent space, bringing it
to a level comparable to that of a standard VAE. It is worth noting that the
global disentanglement may no longer be guaranteed. Rather, the model only
provides local disentanglement within the proximity of each cluster.

Upon training the GridVAE with a 40-dimensional latent space by using the
proposed Eq.~\eqref{eq:vae_kl} on the CelebA dataset, we observe some
intriguing
disentanglement phenomena. Fig.~\ref{fig:vae_unsupervised} showcases the
disentanglement of two latent space dimensions, where the first dimension
governs one attribute and the second dimension determines another one.
Combining
these two dimensions leads to simultaneous attribute changes in the generated
images.
\begin{subfigure}[!ht]
\centering
\setcounter{subfigure}{0}
\begin{minipage}[b]{0.49\textwidth}
    \centering
    \includegraphics[width=1.0\textwidth]{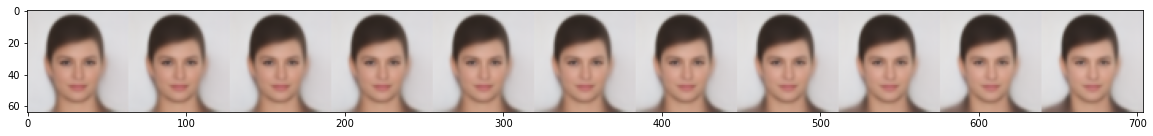}
    +
    \includegraphics[width=1.0\textwidth]{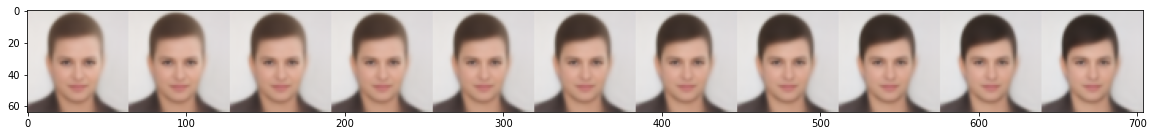}
    =
    \includegraphics[width=1.0\textwidth]{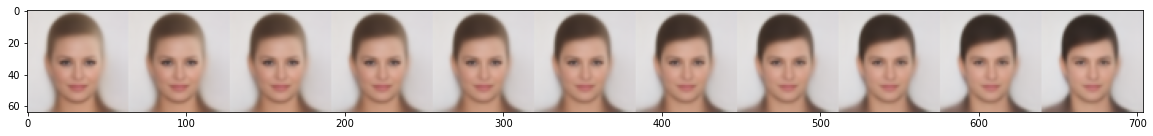}
\end{minipage}
\begin{minipage}[b]{0.49\textwidth}
    \centering
    \includegraphics[width=1.0\textwidth]{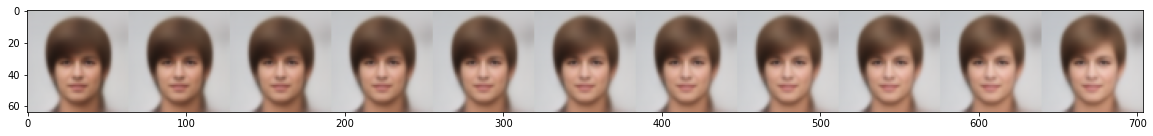}
    +
    \includegraphics[width=1.0\textwidth]{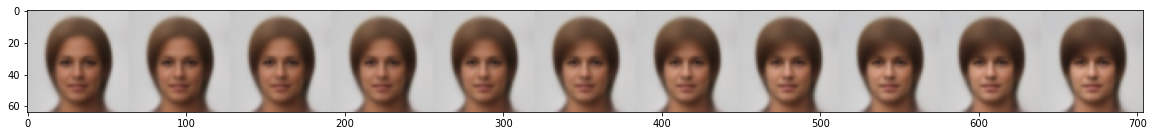}
    =
    \includegraphics[width=1.0\textwidth]{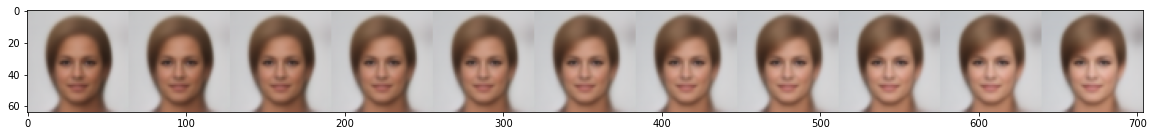}
\end{minipage}
\setcounter{subfigure}{-1}
\itcaption{Two generated examples using linear sampling in the latent space, \textbf{(A)} left 3 rows and \textbf{(B)} right 3 rows.
The top row fixes the dimensions and changes the first one,
\revision{collar in \textbf{(A)} and skin color in \textbf{(B)},}
from -0.5 to +1.5.
The middle row fixes the dimensions and changes the second one,
\revision{hair color in \textbf{(A)} and hairstyle in \textbf{(B)},}
from -0.5 to
+1.5.
The bottom row changes the first and second dimensions from -0.5 to +1.5.}
\label{fig:vae_unsupervised}
\end{subfigure}

An inherent limitation of this unsupervised approach is that while the latent
space appears to be locally disentangled for each image, the same dimension
may
have different semantic interpretations across different images. To address
this
issue, we introduce all 40 attributes of the dataset during the training. This
should establish an upper bound on the disentanglement.

\subsubsection{From Unsupervised to Guided and Partially Guided GridVAE}
To this end, we described an unsupervised approach to learning the latent
space
representation of images. However, for datasets like CelebA with ground truth
attributes, we can incorporate them into the latent space to guide the
learning.
Specifically, we extract the 40-dimensional attribute vector indicating the
presence or absence of each feature for each image in a batch and treat it as
the ground truth cluster centre $\mu_{i}^{gt}$. Hence, instead of rounding the
latent space representation $\mu_i$ in Eq.~\eqref{eq:vae_kl}, we replace it
with
$\mu_{i}^{gt}$.

One limitation of this approach is the requirement of the ground truth
attributes for all images, which may not always be available or feasible.
Additionally, it is important to note that while we refer to this approach as
``guided", the given attribute information only serves in the latent space as
the cluster assignment prior, and the VAE reconstruction task remains
unsupervised. This differs from classical supervised learning, where the label
information is the output. Furthermore, in our approach, no specific
coordinate
in the latent space is designated for the input. Instead, we provide guidance
that the sample belongs to a cluster centred at a certain point in the latent
space.

This guided learning framework can be extended to a subset of the 40
attributes
or a latent space with more dimensions. For clarity, we will refer to the
latter
as ``partially guided" to distinguish it from the commonly used
``semi-supervised" by using a subset of the labelled dataset.

We conduct the experiments using attribute information as latent space priors
and obtain the following findings for the guided approach: (a) GridVAE is able
to cluster images accurately based on their attributes and the same dimension
has the same semantic meaning across different images. For instance, dimension
31 represents ``smile". (b) GridVAE could not generate images for clusters
that
have little or no representation in the training set. For example, the attempt
to generate an image of a bald female by constraining GridVAE to the
``female"
and ``bald" clusters is not achievable for an accurate representation. (c)
Some
attributes are more universal across different images, such as their ability
to
add a smile to almost any face. However, other attributes, such as gender, are
not always modifiable. This could be caused by attributes that are not
independent and can be influenced by others. Universal attributes, such as
``smile", seem to primarily located locally in the image region without
interruption from the other attributes, see Fig.~\ref{fig:vae_smile_female}.
\begin{subfigure}[!ht]
\setcounter{figure}{13}
\centering
\setcounter{subfigure}{0}
\begin{minipage}[b]{0.49\textwidth}
    \includegraphics[width=\linewidth]{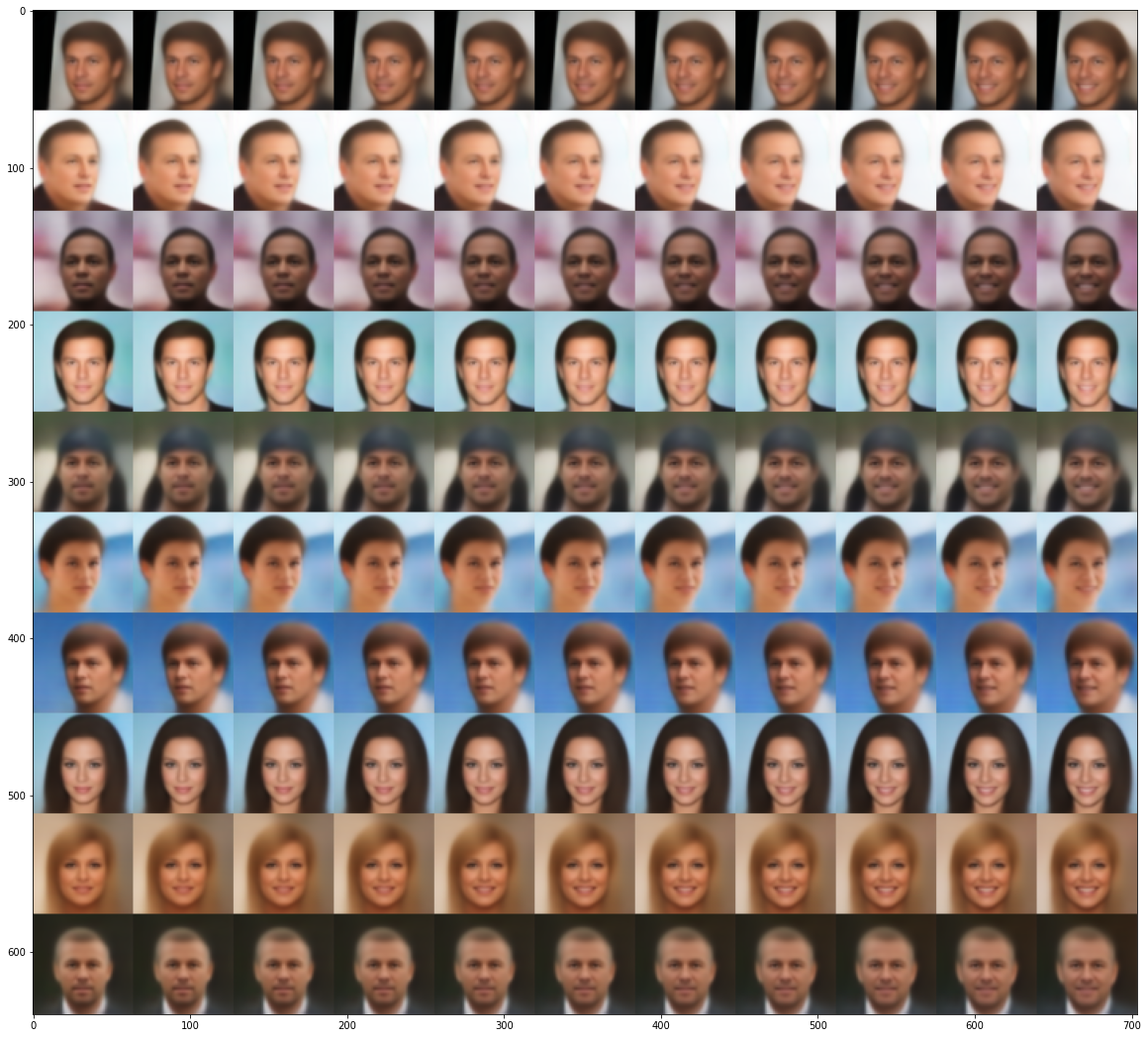}
    \label{fig:vae:smile}
\end{minipage}
\begin{minipage}[b]{0.49\textwidth}
    \includegraphics[width=\linewidth]{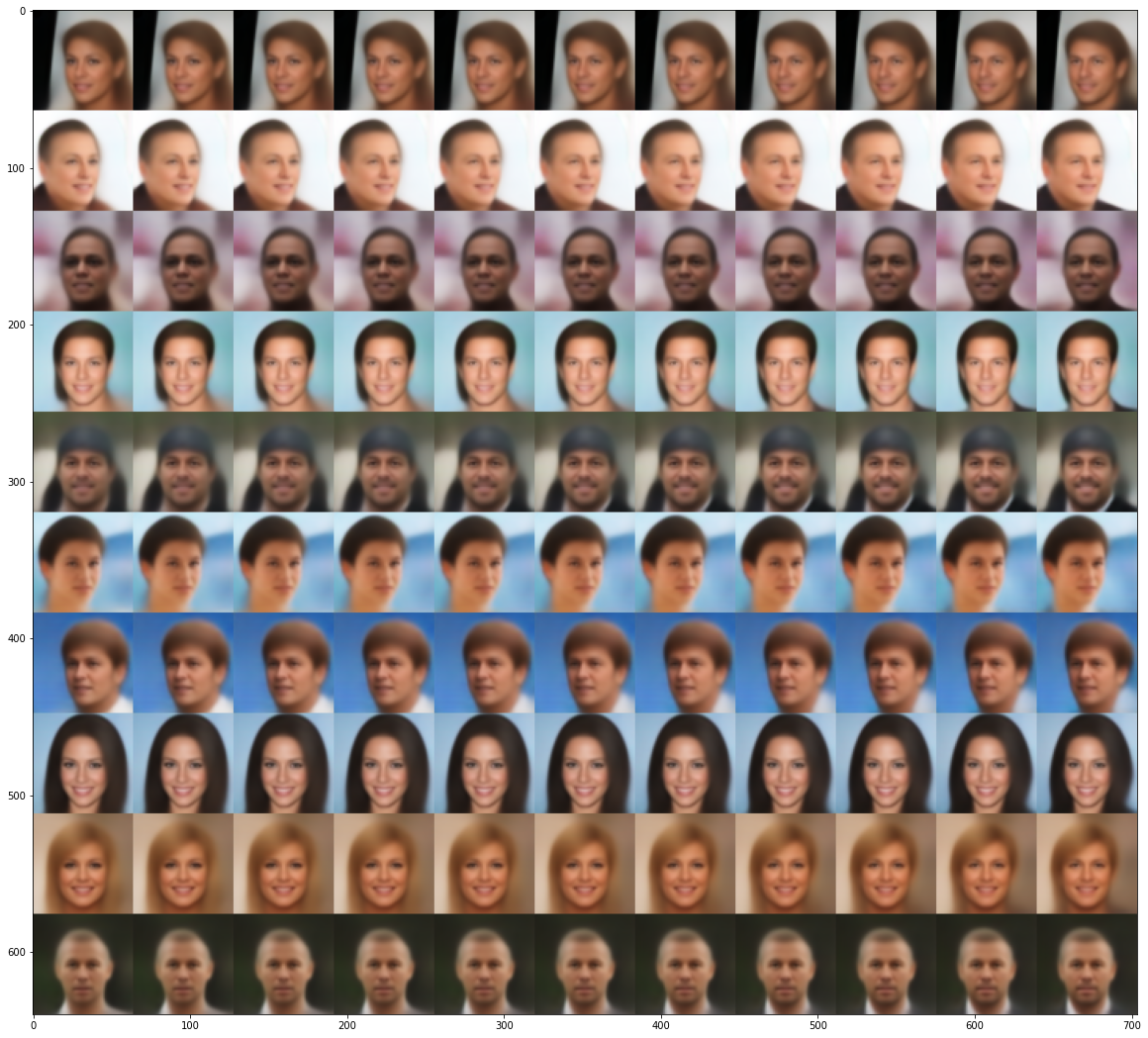}
    \label{fig:vae:female}
\end{minipage}
\setcounter{subfigure}{-1}
\itcaption{Generated images from sampling in the latent space, \textbf{(A)} left 10 rows and \textbf{(B)} right 10 rows. Keeping all
other
dimensions fixed and changing dimension \textbf{(A)} 31 (smile) from -0.5 to
+1.5, or \textbf{(B)} 20 (male) from -0.5 to +1.5.}
\label{fig:vae_smile_female}
\end{subfigure}

\begin{subfigure}[!ht]
\setcounter{figure}{14}
\centering
\setcounter{subfigure}{0}
\begin{minipage}[b]{0.49\textwidth}
    \includegraphics[width=\linewidth]{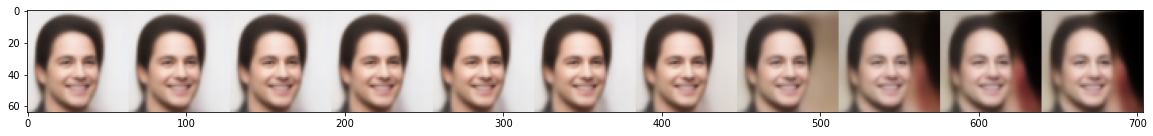}
\end{minipage}
\begin{minipage}[b]{0.49\textwidth}
    \includegraphics[width=\linewidth]{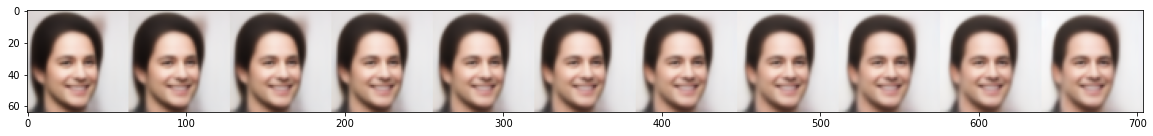}
\end{minipage}
\setcounter{subfigure}{-1}
\itcaption{Partially guided GridVAE generation from the latent attributes
which
are not provided during training. The left and right
\revision{subfigures (each with 11 images)} are with the
dimensions 20 and 31 respectively.}
\label{fig:vae:dim_not_given}
\end{subfigure}
To further illustrate the incompleteness and correlation among the attributes
in
the CelebA dataset, we use a subset of the given attributes. We choose 38 out
of
the 40 attributes, excluding attributes 20 (female/male) and 31 (not
smiling/smiling). Fig.~\ref{fig:vae:dim_not_given} shows that the GridVAE
cannot
learn the omitted attributes. This highlights the interdependence of different
attributes in the latent space.

\subsection{Combining Manifolds of GridVAE Disentangled Attribute and Facial
Recognition}
After achieving a disentangled latent space, one may still wonder about the
usefulness of a semantic description of a manifold. One can consider the
scenario where another manifold, such as a facial recognition manifold, is
learned. By studying these two manifolds jointly, we can gain insights to make
the models more explainable and useful. One potential application is to better
understand the relationship between facial attributes and facial recognition.
By
analyzing the disentangled latent space of facial attributes and the manifold
learned for facial recognition, we can potentially identify which attributes
are
the most important for recognising different faces. This understanding can
then
be used to improve the performance of facial recognition models as well as
explain the model decisions.

For instance, FaceNet~\citep{schroff2015facenet} directly learns a mapping
from
face images to a compact Euclidean space where distances correspond to a
measure
of face similarity. To discover the semantic structure of this  manifold with
$\mb{x}$ as binary attributes, we can follow these steps:
\begin{enumerate} 
\item Build a face recognition manifold using contrastive learning. 
\item Use the CelebA dataset with ground truth attribute labels (40 binary
values). 
\item Insert CelebA samples onto the recognition manifold. 
\item Find the nearest neighbour for each CelebA sample using the face
recognition manifold coordinates. 
\item For each attribute in $\mb{x}$, compute $p(\mb{x})$ over the entire CelebA
dataset.
\item For each attribute in  $\mb{x}$, compute $p(\mb{x}| \mb{x} \text{ of nearest neighbour}
=
0)$. 
\item For each attribute in $\mb{x}$, compute the KL divergence between $p(\mb{x})$ and
$p(\mb{x}| \mb{x} \text{ of nearest neighbour} = 0)$. 
\item Identify attributes with the largest KL divergence. \end{enumerate}

Fig.~\ref{fig:vae:2_manifolds} demonstrates that the KL divergence between
$p(\mb{x})$ and $p(\mb{x}| \mb{x} \text{ of nearest neighbour} = 0)$ is significantly larger
for certain attributes, such as ``male", ``wearing lipstick", ``young" and
``no
beard", than the others. This indicates that the neighbourhood structure of
the
facial recognition manifold is markedly different from the distribution of
these
attributes in the entire dataset. These findings highlight the importance of
the
joint study of different manifolds to gain a more profound understanding of
the
relationship between the attributes and the recognition tasks. By
incorporating
it into the models, we can potentially improve the performance of facial
recognition models and also enhance their interpretability.
\begin{subfigure}[t]
\setcounter{figure}{15}
\centering
\setcounter{subfigure}{0}
\begin{minipage}[b]{0.66\textwidth}
    \includegraphics[width=\linewidth]{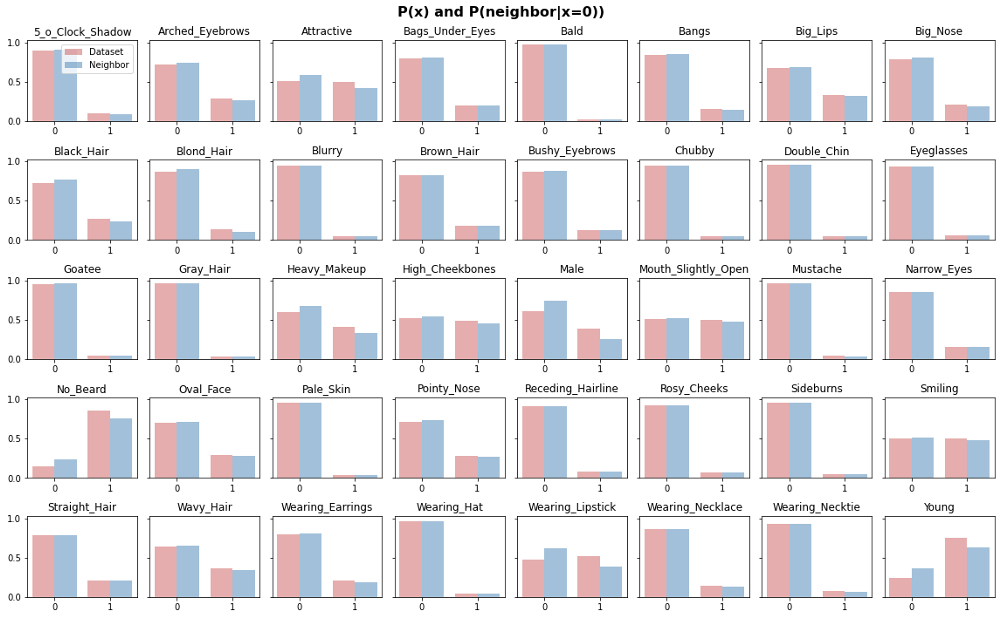}
    \itcaption{$p(x)$  and $p(x| x \text{ of nearest neighbour} = 0)$
distributions.}
\end{minipage}
\begin{minipage}[b]{0.33\textwidth}
    \includegraphics[width=\linewidth]{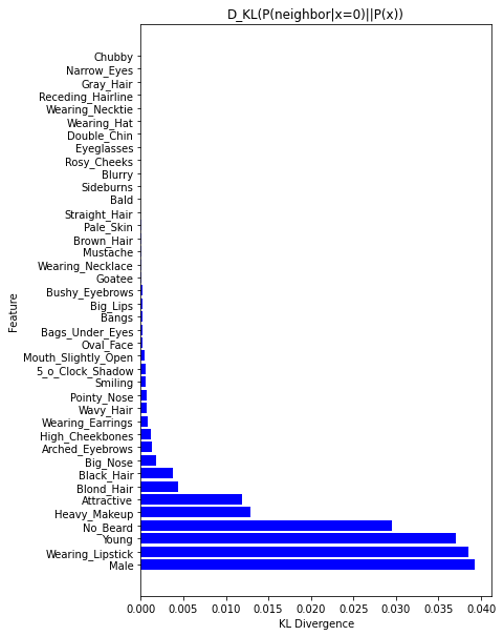}
    \itcaption{KL divergence.}
\end{minipage}
\setcounter{subfigure}{-1}
\vspace{-5mm}
\itcaption{Semantic structure of the face recognition manifold by jointly
studying the attribute manifold and the facial recognition manifold.}
\label{fig:vae:2_manifolds}
\end{subfigure}

%
\section{Application to Defend Patch Attacks}
\label{sec:attacks}

\revision{To this end, interpretable and controllable samplings from
each semantic distribution on the manifold can be achieved by using the
semantic disentanglement in Section~\ref{sec:el} towards high-fidelity
and diverse image generation and probability distribution analysis in
Section~\ref{sec:mle}.
It is also of strong interest to enhance the robustness of such semantic
samplings under certain attacks.
In this section, we present an adversarial robustness framework by
enforcing the semantic consistency between the classifier and the decoder
for reliable density estimation on the manifold.}

\revision{\subsection{Adversarial Defence with Variational Inference}

In \citep{yang2022adaptive}, adversarial robustness can be achieved
by enforcing the semantic consistency between a decoder and a
classifier (adversarial robustness does not exist in non-semantically
consistent classifier-decoder). We briefly review the adversarial
purification framework below. We define the real-world high-dimensional
data as
$\mb{x} \in \mathbb{R}^n$ which lies on a low-dimensional manifold
$\mathcal{M}$ diffeomorphic to $\mathbb{R}^m$ with $m \ll n$. We define an
encoder function ${f}:\mathbb{R}^n \to \mathbb{R}^m$ and a decoder
function $f^{\dagger}:\mathbb{R}^m \to \mathbb{R}^n$ to form an
autoencoder. For a point $\mb{x} \in \mathcal{M}$, ${f}^{\dagger}$
and ${f}$ are approximate inverses. We define a discrete label set
$\mathcal{L}$ of $c$ elements as $\mathcal{L}=\{1,...,c\}$ and a classifier
in
the latent space as ${h}: \mathbb{R}^m \to \mathcal{L}$. The encoder
maps
the image $\mb{x}$ to a lower-dimensional vector $\mb{z} =
{f}(\mb{x}) \in \mathbb{R}^m$ and the functions ${f}$ and
${h}$ together form a classifier in the image space
${h}(\mb{z}) = ({h} \circ {f})(\mb{x}) \in
\mathcal{L}$. 

A classifier (on the manifold) is a semantically consistent classifier if
its predictions are consistent with the semantic interpretations of the
images reconstructed by the decoder. Despite that the classifiers and
decoders (on the manifold) have a low input dimension, it is still
difficult to achieve high semantic consistency between them. Thus,
we assume that predictions and reconstructions from high data density
regions of $p(\mathbf{z}|\mathbf{x})$ are more likely to be semantically
consistent and we need to estimate the probability density in the latent
space with the variational inference. 

We define three sets of parameters: 1) $\phi$ parametrises the encoder
distribution, denoted as $q_{\phi}(\mb{z} | \mb{x})$, 2) $\theta$
parametrises the decoder distributions, represented as $p_{\theta}(\mb{x}
|
\mb{z})$, and 3) $\psi$ parametrises the classification head, given by
${h}_{{\psi}}(\mb{z})$.  These parameters are jointly optimised
with respect to the ELBO loss and the cross-entropy loss as shown in
Eq.~\eqref{eq:joint_elbo_cls_kkt}, where $\lambda$ is the trade-off term
between
the ELBO and the classification loss.
We provide the framework in Figs.~\ref{fig:defence_flow_chart}\textbf{(A)}-\textbf{(B)} for the two-stage procedure and 
the trajectory of cluster center change after introducing our purification over attacks in Fig.~\ref{fig:defence_flow_chart}\textbf{(C)}.
By adopting this formulation, we notice
a
remarkable semantic consistency between the decoder and the classifier.
Specifically, on Fashion-MNIST~\citep{fashion-mnist}, when making predictions on adversarial examples, if the
predicted
label is ``bag", we observe that the reconstructed image tends to resemble a
``bag" as well. This phenomenon is illustrated in
Fig.~\ref{fig:defence_flow_chart}\textbf{(D)} and Fig.~\ref{fig:puridied}.


\begin{figure}[!t]
\centering
\includegraphics[width=0.9\textwidth]{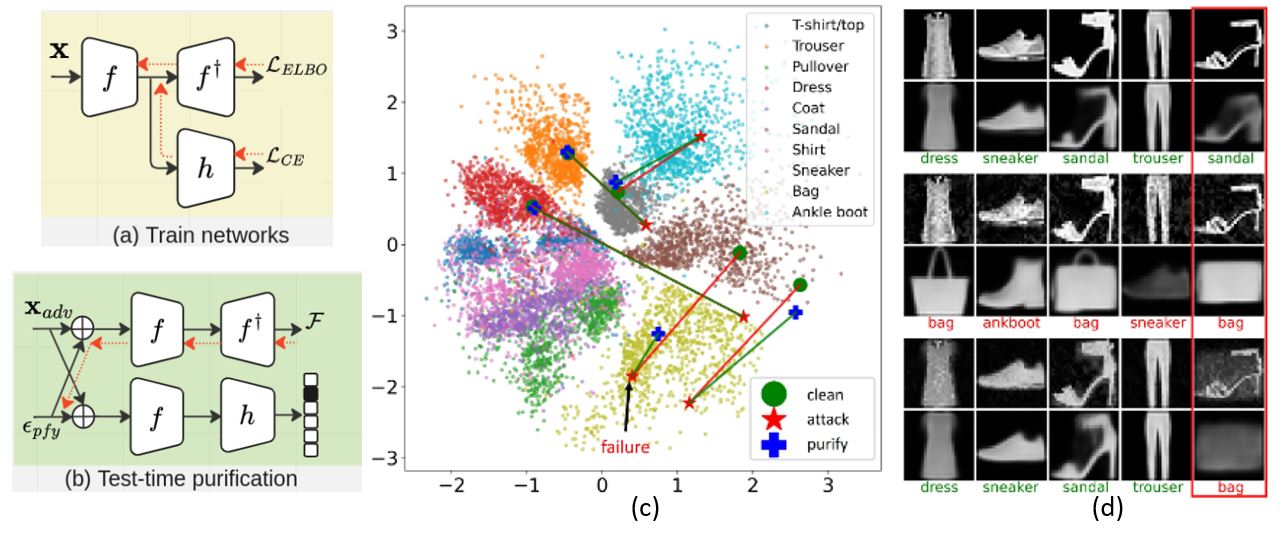}
\itcaption{The framework of adversarial purification for image-level
adversarial
attacks. \textbf{(A)} Jointly train the classifier with the ELBO loss.
\textbf{(B)} Test time adversarial purification with the ELBO loss.
\textbf{(C)}
Trajectories of clean (green) - attack (red) - purified (blue) images on a 2D
latent space. \textbf{(D)} Input images and reconstruction images of samples
in
\textbf{(C)}. The top two rows are the input and reconstruction of clean
images,
the middle two rows are the input and reconstruction of adversarial images.
The
bottom two rows are the input and reconstruction of purified images. The text
represents predicted classes with {\color{ForestGreen}green} colour for
correct
predictions and {\color{red}red} colour for incorrect predictions. The red box
on the right corresponds to the failure case (purified process fails). }
\label{fig:defence_flow_chart}
\vspace{-3mm}
\end{figure}

\begin{equation}
\max_{{\theta} ,{\phi}, {\psi}} \underbrace{\mathbb{E}_{\mb{z}
\sim
q_{{\phi}}(\mb{z} | \mb{x} )}{\left[\log p_{{\theta}}(\mb{x}
|
\mb{z})\right]}   - D_{KL}[q_{{\phi}}(\mb{z} | \mb{x} ) \|
p(\mb{z})]}_{\text{ELBO (lower bound of} \log p_{{\theta}}(\mb{x})
\text{)}} + \lambda \underbrace{\mathbb{E}_{\mb{z} \sim
q_{{\phi}}(\mb{z} | \mb{x} )}[\mb{y}^\intercal \log
{h}_{{\psi}}(\mb{z})]}_{\text{Classification loss}}\ .
\label{eq:joint_elbo_cls_kkt}
\end{equation}

To defend against image-level attacks, a purification vector can be obtained through
the test-time optimisation over the ELBO loss. For example, given an
adversarial
example $\mb{x}_{\rm adv}$, a purified sample can be obtained by
$\mb{x}_{\rm pfy} = \mb{x}_{\rm adv} + \bm{\epsilon}_{\rm pfy}$ with
\begin{equation}
\label{eq:purify_jointly_obj}
\begin{aligned}
\bm{\epsilon}_{\rm pfy} = \argmax_{\bm{\epsilon} \in \mathcal{C}_{\rm pfy}}
\mathbb{E}_{\mb{z} \sim q_{{\phi}}(\mb{z} | \mb{x}_{\rm adv} +
\bm{\epsilon})}{\left[\log p_{{\theta}}(\mb{x}_{\rm adv} +
\bm{\epsilon}|
\mb{z})\right]} - D_{\rm KL}[q_{{\phi}}(\mb{z} | \mb{x}_{\rm
adv}
+ \bm{\epsilon}) \| p(\mb{z})]\ ,
\end{aligned}
\end{equation}
where $\mathcal{C}_{\rm pfy}=\{\bm{\epsilon} \in \mathbb{R}^n \mid
\mb{x}_{\rm adv} + \bm{\epsilon} \in [0,1]^n \text{ and }
\|\bm{\epsilon}\|_{p} \leq \epsilon_{\rm th} \}$ which is the feasible set for
purification and $\epsilon_{\rm th}$ is the purification budget.
Since the classifier and the decoder are semantically consistent, the
predictions from the classifier become normal to defend against the attacks
upon normal reconstructions.}

\subsection{Bounded Patch Attack}
In this work, we focus on the $\ell_0$-bounded attacks~
\citep{patch_adv,DBLP:conf/eurosp/PapernotMJFCS16}
from the manifold perspective which is not investigated in the prior work.
In contrast to
full image-level attacks like $\ell_2$ and $\ell_\infty$
bounded
attacks \citep{PGD_attack}, patch attacks, which are $\ell_0$ bounded
attacks, aim to restrict
the
number of perturbed pixels. These attacks are more feasible to implement in
real-world settings, resulting in border impacts. Below, we conduct an initial
investigation into the defence against patch attacks by leveraging the
knowledge
of the data manifold.

When compared to $\ell_\infty$ attacks, $\ell_0$ attacks, such as the
adversarial patch attacks, introduce larger perturbations to the perturbed
pixels. Therefore, we decide to remove the purification bound for the
patch-attack purification. Without these
constraints, the purified examples can take on any values within the image
space. A purification vector can then be obtained through
the test-time optimisation over the ELBO loss as shown in Eq.~\eqref{eq:purify_jointly_obj}.

\subsection{Experiments}

\begin{figure}[!t]
\centering
\includegraphics[width=0.7\textwidth]{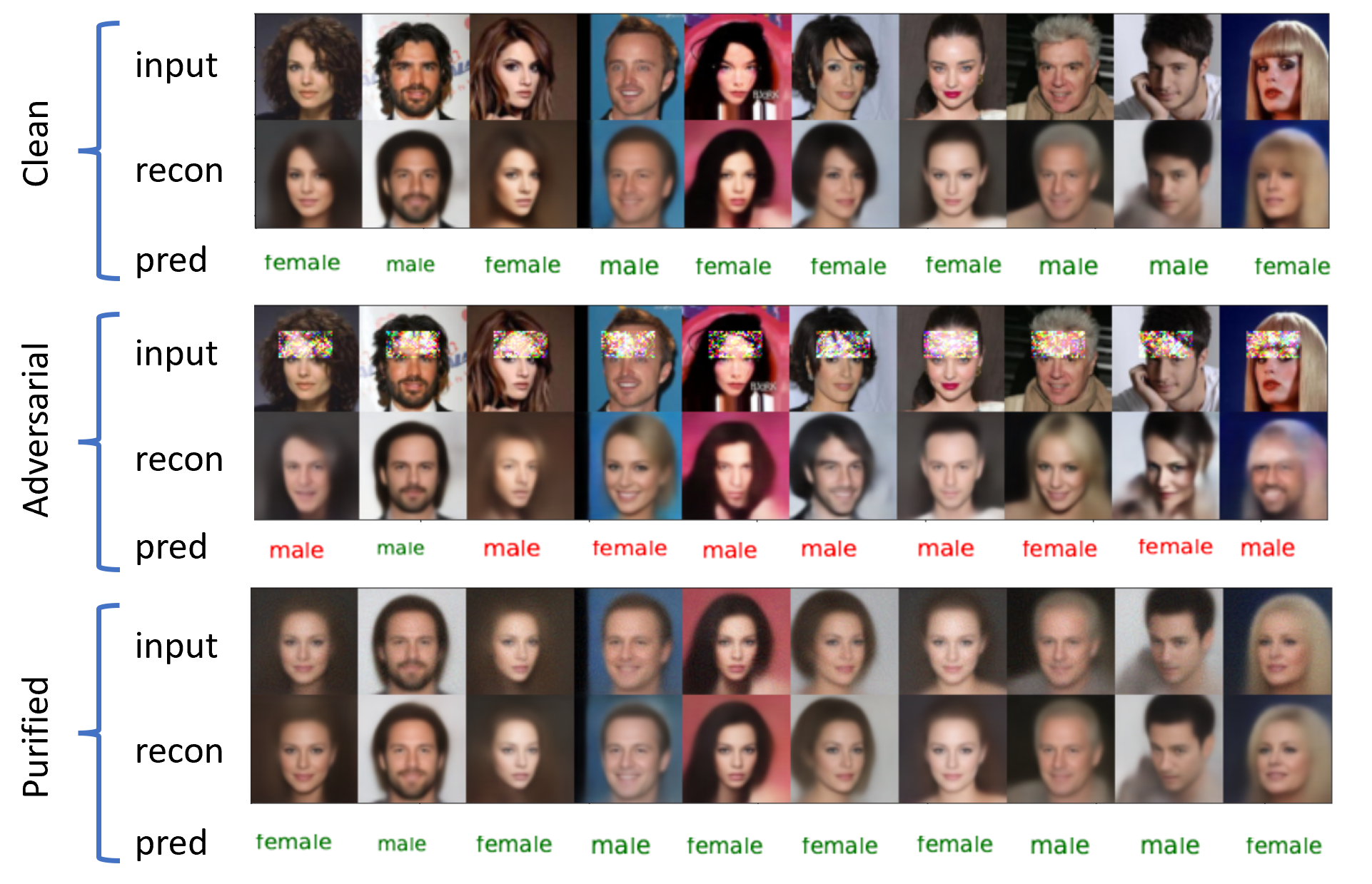}
\itcaption{Class predictions from the VAE-Classifier models on clean,
adversarial and purified samples of the CelebA gender attribute. The top two
rows are the input and reconstruction of clean images, the middle two rows are
the input and reconstruction of patch adversarial images. The bottom two rows
are the input and reconstruction of purified images. The text represents the
predicted classes with {\color{ForestGreen}green} colour for correct
predictions
and {\color{red}red} colour for incorrect predictions. Since predictions and
reconstructions from the VAE classifier are correlated, our test-time defences
are effective against adversarial attacks.}
\label{fig:puridied}
\vspace{-3mm}
\end{figure}

\begin{table}[!ht]
    \centering
    \itcaption{Classification accuracy of the model on clean and adversarial
(patch) examples.}
        \resizebox{0.85\linewidth}{!}{
        \label{tbl:celebA}
        \begin{tabular}{l|ccc|ccc}
            \hline
            & \multicolumn{3}{c|}{VAE-CLF} & \multicolumn{3}{c}{+TTD (ELBO)}
\\
            Dataset (Backbone) & Clean &  Patch-PGD  &  Patch-NAG & Clean & 
Patch-PGD  & Patch-NAG \\
            \hline
            CelebA-Gender (ResNet-50) & 97.86 & 13.14 & 6.83  & 91.20 &
{75.75} & {76.75} \\
            \hline
        \end{tabular}
    }
    \label{tbl:rev_performacne}
\end{table}
We use the gender classification model \citep{yang2022adaptive} to demonstrate
the adversarial purification of $\ell_0$ bounded attacks. To ensure that the
adversarial examples do not alter the semantic content of the images, we
restrict the perturbation region to the forehead of a human face. The patch
for
perturbation is a rectangular shape measuring $16\times32$, see
Fig.~\ref{fig:puridied}. For the patch attacks, we conduct 2,048 iterations
with
step size $1/255$ using PGD \citep{PGD_attack} and PGD-NAG (Nesterov
Accelerated
Gradient) \citep{LinS00H20}. In
Table~\ref{tbl:rev_performacne},
the purification is carried out through 256 iterations with the same step
size.

%
\revision{\section{Limitation}
\label{sec:limitation}
The current version of log-probability estimation in diffusion models has
limitations in evaluating high-dimensional images.
Specifically, at early denoising steps (when $t$ is small) the diffusion model
serves as a denoiser such that $\mb{x}_t$ and $\mb{x}_{t+h}$ are
similar while at large steps (when $t$ moves towards $T$), their difference
is still small due to the high proportion of the Gaussian noise in $\mb{x}_t$.
This leads to the proportion of the difference between $\mb{x}_t$ and
$\mb{x}_{t+h}$ for effective out-of-distribution detection small
compared with the $\log p$ accumulated in the processes.
We keep this as an open problem for future work.}

%
\section{Conclusion}
\label{sec:conclusion}
This work studies the image geometric representation from high-dimensional
spatial space to low-dimensional latent space on the image manifold. To explore
the
image probability distribution with the assumption that real images are
usually
in a high-density region while not all samples from the distribution can be
represented as realistic images, we incorporate log-likelihood estimation
into the procedures of normalising flows and diffusion models.
\revision{Meanwhile, we explore the hierarchical normalising flow structure
and a higher-order solution in diffusion models for high-quality and
high-fidelity image generation.
For an interpretable and controllable sampling from the semantic distribution
on the manifold, 
we then propose GridVAE model under an EL framework to disentangle the
elements of the latent variable
on the image manifold.
To test the semantic and reconstruction robustness on the manifold, we first
apply patch attacks and defences in the image space and then effectively
recover the semantics under such attacks with our purification loss.}
Experiments show the effectiveness of probability
estimation in distinguishing seen examples from unseen ones, the quality and the
efficiency with large sampling steps in image generation, \revision{meaningful
representations of
varying specific element(s) of the latent variable to control the object
attribute(s) in the image space, and the well-preserved
semantic consistency with patch attacks.}

%
\section*{Conflict of Interest Statement}
Authors Peter Tu, Zhaoyuan Yang, and Yiwei Fu are employed by General Electric.
The remaining authors declare that the research was conducted in the absence of
any commercial or financial relationships that could be construed as a potential
conflict of interest.

%
\section*{Author Contributions}
Peter Tu and Richard Hartley are the principal investigators of this project.
Zhiwei Xu,
Richard Hartley, Jing Zhang, and Dylan Campbell contribute to the likelihood
estimation section;
Zhaoyuan Yang and Peter Tu contribute to the attacks and defences section;
and Yiwei Fu
contributes to the semantic disentanglement section. All authors contribute to
discussions and proofreading in this work.

%
\section*{Funding}
This work is supported by the DARPA geometries of
learning (GoL) project under the grant agreement number HR00112290075.

%
\section*{Acknowledgments}
We thank Amir Rahimi for his contribution to the code and discussion of the
normalising flow models.

%
\section*{Supplemental Data}
\ifbool{arxiv}
{Please refer to the appendix.}
{Please refer to ``Supplementary Material".}

%
\section*{Data Availability Statement}
Datasets used in this work \revision{include MNIST~\citep{mnist},
Fashion-MNIST~\citep{fashion-mnist}, CelebA~\citep{liu2015faceattributes},
CIFAR10~\citep{cifar10}, and Point Set (swiss roll, moon,
``S", etc.)~\citep{scikit-learn},} and are available online.
\revision{Data split for training and validation / testing follows the
standard scripts provided in the references; for Point Set data, we randomly
sampled 10,000 points for both seen and unseen shapes.}
Our code with demonstrated samples and \revision{hyperparameters for learning}
will be released upon publication.
For further inquiries, please contact the corresponding author.

\ifbool{arxiv}{
\clearpage
\setcounter{section}{0}
\section*{Appendix}

\section{The Runge--Kutta Method for Diffusion Models}
We first revisit the Runge--Kutta method (RK4) which solves initial
value problems~\citep{RK41,RK42,RK4Wiki}. Given an
initial value problem as $f(\mb{x}_t, t)=d\mb{x}_t / dt$, where
$\mb{x}_t$ is associated with time $t$, the estimation of $\mb{x}_t$ at
time $(t + h)$ with step size $h$ is computed by
\begin{equation}
\label{eq:rk4_iter}
    \mb{x}_{t+h}
    = \mb{x}_t + \frac{h}{6}
    \left( \mb{k}_1 + 2 \mb{k}_2 + 2 \mb{k}_3 + \mb{k}_4 \right)\ ,
\end{equation}
where
\begin{alignat}{2}
    \mb{k}_1 &= f (\mb{x}_t, t )\ ,\\
    \mb{k}_2 &= f ( \mb{x}_t + \frac{h}{2} \mb{k}_1&&, t + \frac{h}{2} )\ ,\\
    \mb{k}_3 &= f ( \mb{x}_t + \frac{h}{2} \mb{k}_2&&, t + \frac{h}{2} )\ ,\\
    \mb{k}_4 &= f ( \mb{x}_t + h \mb{k}_3&&, t + h )\ .
\end{alignat}

For an initial value $\mb{x}_0$ at time $(t=0)$, one can estimate $\mb{x}_T$ iteratively
by using Eq.~\eqref{eq:rk4_iter} from $(t=0)$ to
the terminate time $(t=T)$.

For the backward process of a diffusion model,
the reverse step can be written as
\begin{equation}
    \mb{x}_{t- h} = g \left( \mb{x}_t, \boldsymbol{\epsilon}_{t - h}, t - h \right)\ ,
\end{equation}
where $g(\cdot)$ refers to the backward step in DDPM~\citep{ddpm} with Gaussian noises
or DDIM~\citep{ddim} without Gaussian noises and
$\boldsymbol{\epsilon}_{t - h}$ is the reversing gradient 
from $\mb{x}_{t}$ to $\mb{x}_{t - h}$, denoted as ``model prediction".
To apply RK4 to the reverse step, we follow the same rule
as Eq.~\eqref{eq:rk4_iter} but change
the moving step as
\begin{equation}
\label{eq:diff_iter}
    \boldsymbol{\epsilon}_{t - h}
    = \frac{1}{6} \left( \mb{k}_1 + 2 \mb{k}_2 + 2 \mb{k}_3 + \mb{k}_4 \right)\ ,
\end{equation}
where
\begin{alignat}{2}
    \mb{k}_1 &= f (\mb{x}_t, t )\ ,\\
    \mb{k}_2 &= f ( g (\mb{x}_t, \mb{k}_1&&, t - \lbound{ \frac{h}{2} } ), t - \lbound{ \frac{h}{2} } )\ ,\\
    \mb{k}_3 &= f ( g (\mb{x}_t, \mb{k}_2&&, t - \lbound{ \frac{h}{2}} ), t - \lbound{ \frac{h}{2}} )\ ,\\
    \mb{k}_4 &= f ( g (\mb{x}_t, \mb{k}_3&&, t - h ), t - h )\ ,
\end{alignat}
where $\lbound{\cdot}$ rounds to the smaller integer due to the discretization of
the sampling time space.
The difference of moving steps between Eq.~\eqref{eq:diff_iter} and Eq.~\eqref{eq:rk4_iter}
is the multiplication of $h$ or $(h / 2)$ to $\mb{k}_i$
for $i=\{1,2,3,4\}$.
Empirically, applying these multiplications in diffusion sampling leads to
strongly unrealistic image generation even with small $h$.
We thus hypothesize that the moving steps are already considered in $g(\cdot)$ because of
the joint effect of the sampling coefficients in $g(\cdot)$, $\boldsymbol{\epsilon}_{t}$,
$\boldsymbol{\epsilon}_{t-\lbound{ h / 2 }}$, and $\boldsymbol{\epsilon}_{t-h}$.

\clearpage
}

\bibliographystyle{references}
\bibliography{references}

\end{document}